%% file: main.tex
\documentclass[11pt]{article}

\usepackage[preprint]{acl}

\usepackage{times}
\usepackage{latexsym}

\usepackage[T1]{fontenc}

\usepackage[utf8]{inputenc}

\usepackage{microtype}

\usepackage{inconsolata}

\usepackage{graphicx}

\usepackage{float}

\usepackage{verbatim}

\usepackage[table]{xcolor}
\usepackage{colortbl}
\usepackage{multirow}

\definecolor{lowcolor}{RGB}{68,1,84}     %
\definecolor{midcolor}{RGB}{58,162,173}  %
\definecolor{highcolor}{RGB}{253,231,37} %
\usepackage{array}
\usepackage{amsmath}
\definecolor{redcell}{RGB}{255,102,102}
\definecolor{greencell}{RGB}{102,255,102}
\definecolor{yellowcell}{RGB}{255,255,153}
\definecolor{orangecell}{RGB}{255,204,102}

\title{Unveiling Intrinsic Dimension of Texts: from Academic Abstract to Creative Story}

\author{
  \textbf{Vladislav Pedashenko\textsuperscript{1*}},
  \textbf{Laida Kushnareva\textsuperscript{2*}},
  \textbf{Yana Khassan Nibal\textsuperscript{2}},
  \textbf{Eduard Tulchinskii\textsuperscript{2}},\\
  \textbf{Kristian Kuznetsov\textsuperscript{2}},
  \textbf{Vladislav Zharchinskii\textsuperscript{1}},
  \textbf{Yury Maximov\textsuperscript{3}},
  \textbf{Irina Piontkovskaya\textsuperscript{2}}
  \\
  \\
  \textsuperscript{*}Equal Contribution,\\
  \\
  \textsuperscript{1} Moscow State University,
  \textsuperscript{2} Lomonosov Research Institute
  \textsuperscript{3} Interdata Astana
  }

\input{math_commands.tex}

\usepackage{hyperref}
\usepackage{url}

\usepackage{amsmath,amssymb,amsfonts}
\usepackage{bm}
\usepackage{hyperref}
\usepackage{mathtools}
\usepackage[inline]{enumitem}
\usepackage{booktabs}

\newcommand{\xvec}{\mathbf{x}}
\newcommand{\xh}{\hat{\mathbf{x}}}
\newcommand{\fvec}{\mathbf{f}}
\newcommand{\Wenc}{{\mathbf{W}}_{\mathrm{enc}}}
\newcommand{\Wdec}{{\mathbf{W}}_{\mathrm{dec}}}
\newcommand{\benc}{{\mathbf{b}}_{\mathrm{enc}}}
\newcommand{\bdec}{{\mathbf{b}}_{\mathrm{dec}}}

\newcommand{\dvec}{\mathbf{d}}

\begin{document}
\maketitle
\begin{abstract}
Intrinsic dimension (ID) is an important tool in modern LLM analysis, informing studies of training dynamics, scaling behavior, and dataset structure, yet its textual determinants remain underexplored. We provide the first comprehensive study grounding ID in interpretable text properties through cross-encoder analysis, linguistic features, and sparse autoencoders (SAEs). In this work, we establish three key findings. First, ID is complementary to entropy-based metrics: after controlling for length, the two are uncorrelated, with ID capturing geometric complexity orthogonal to prediction quality. Second, ID exhibits robust genre stratification: scientific prose shows low ID ($\sim$8), encyclopedic content medium ID ($\sim$9), and creative/opinion writing high ID ($\sim$10.5) across all models tested. This reveals that contemporary LLMs find scientific text "representationally simple" while fiction requires additional degrees of freedom. Third, using SAEs, we identify causal features: scientific signals (formal tone, report templates, statistics) reduce ID; humanized signals (personalization, emotion, narrative) increase it. Steering experiments confirm these effects are causal.
Thus, for contemporary models, scientific writing appears comparatively “easy,” whereas fiction, opinion, and affect add representational degrees of freedom. Our multi-faceted analysis provides practical guidance for the proper use of ID and the sound interpretation of ID-based results.

\end{abstract}

\section{Introduction}

Quantifying data complexity is one of the key pillars supporting modern ML and LLM theory. Early information–theoretic views (e.g., the Information Bottleneck) tie learning efficiency to compressibility \cite{tishby2015deep}, while the manifold perspective motivates a geometric view of representations. In NLP, complexity assessments have evolved from data–intrinsic measures (e.g., gzip ratio, lexical or syntactic indicators) to model–based ones that depend on how a model organizes and predicts text. The latter divide into \emph{predictive} metrics (entropy, cross-entropy, perplexity) and \emph{geometric} metrics (anisotropy, intrinsic dimension). While entropy-like measures are ubiquitous in training and evaluation, geometric complexity remains comparatively underexplored.

We focus on \textbf{intrinsic dimension (ID)} as a geometric measure of degrees of freedom in embedding space. We establish a conceptual separation between ID and prediction-based entropy: ID depends solely on the \emph{geometry} of hidden representations, whereas entropy arises only after \emph{unembedding} and reflects alignment with the vocabulary. Hence, the two quantities are not substitutes. We begin by showing empirically that, after controlling for text length, ID and (cross-)entropy are essentially uncorrelated; we then substantiate this separation with a formal argument.

In this paper, we provide a comprehensive, semantics-grounded account of ID for text. First, we characterize how ID varies across domains and styles, showing that scientific/informational prose is consistently low-ID, while opinionated and fictional writing is high-ID, indicating increased representational degrees of freedom from the model’s perspective. Second, we link ID to observable linguistic regularities: higher lexical diversity and lower cross-sentence repetition associate with higher ID, whereas syntactic indicators are comparatively uninformative for well-formed text. Third, using sparse autoencoders and feature steering, we connect ID to interpretable semantic axes (e.g., genre, narrative, personalization, emotional tone) and probe their causal effects on generation. Together, these results position ID as a complementary, geometry-based lens on textual complexity with practical implications for evaluation and training: specifically, the need to balance low-ID (encyclopedic/scientific) and high-ID (personal/narrative) data to faithfully assess model capabilities.

The paper is organized as follows:
Section~\ref{sec:ID_background} formalizes ID and its relation to predictive metrics; Section~\ref{sec:semantic_background} details our SAE-based analysis methodology; Section~\ref{sec:results} presents experimental observations.
Finally, Section~\ref{sec:discussion} summarizes our findings and discusses implications.

\subsection{Related Work}

\paragraph{Intrinsic Dimension in Neural Networks.} Intrinsic dimension (ID) analysis provides a geometric framework for understanding and improving large language models (LLMs). Early work by \citet{li2018measuringintrinsicdimensionobjective} showed that good solutions lie in low-dimensional subspaces of the parameter space. Extending this idea to NLP, \citet{aghajanyan-etal-2021-intrinsic} demonstrated that task-specific variation in pretrained models is confined to a low-dimensional subspace, inspiring low-rank adaptation methods such as LoRA~\cite{hu2022lowrank}, which assume that effective fine-tuning occurs within compact intrinsic subspaces.

Recent studies have shifted from parameter to representation space, using nonlinear estimators to measure ID in embeddings. While \citet{NEURIPS2024_4a68bb0e} link ID to model scaling, \citet{razzhigaev-etal-2024-shape} demonstrate that it may reflect training saturation and dynamics.  \citet{viswanathan2025geometrytokensinternalrepresentations} confirm previous observations and show that higher ID correlates with higher entropy and loss. ID has also been used for architecture search \cite{He_Yao_Wang_Tang_Cheung_See_Han_Chu_2023}, providing an efficient geometric signal for model evaluation. Further, \citet{arnold-2025-memorization} relate ID to memorization capacity, while \citet{lee-etal-2025-geometric} show that nonlinear ID captures semantic compositionality, unlike linear measures such as PCA.

\paragraph{Persistent Homology Dimension (PHD) in Neural Networks.} Intrinsic dimension can be estimated via Persistent Homology Dimension (PHD), which combines local and global properties of point cloud ~\cite{schweinhart2021persistent}. \citet{birdal2021intrinsic} link it to generalization property of neural networks, while~\citet{NEURIPS2023_7baa48bc} show that texts generated by early GPT models exhibit lower PHD than human-written texts, enabling simple AI text detection. However, this distinction diminishes for newer models \cite{kuznetsov-etal-2024-robust}.
Building on this, \citet{kushnareva2024aigeneratedtextboundarydetection} demonstrate that local changes in intrinsic dimensionality can detect boundaries between human and machine-generated text segments.

\paragraph{Other Intrinsic Dimension Approaches.} Several studies explore ID-based analysis for downstream tasks.
\citet{tsukagoshi-sasano-2025-redundancy} show that substantial dimensionality reduction of task-specific embeddings is possible with minimal performance loss across classification, clustering, and retrieval tasks, using estimators such as TwoNN and metrics like Isoscore.
\citet{YinSC24_thuthfullness} apply Local Intrinsic Dimension (LID) to hallucination detection, showing that high and highly variable LID signals fabricated or unsupported content, whereas truthful generations exhibit smoother, low-variance LID profiles.
\citet{ruppik2025morelocalintrinsicdimensions} further argue that intrinsic dimensionality varies locally across embedding space: common tokens occupy simple, low-LID neighborhoods, while rare or domain-specific terms require higher-dimensional representations. Layer-wise analysis reveals that lower layers exhibit uniformly low LID, whereas higher layers show increased LID and variance, reflecting growing specialization and semantic disambiguation.

This rapidly expanding body of research demonstrates that intrinsic dimension in neural networks is not merely a mathematical curiosity, but rather a practical tool that offers pathways toward more interpretable and capable models and opens new applications in downstream tasks.

\section{Intrinsic Dimension}\label{sec:ID_background}

Intrinsic dimension (ID) measures a dataset’s degrees of freedom: for linear spaces and smooth manifolds it matches the usual dimension, and extensions to metric spaces aim to remain robust to noise and recoverable from finite samples.

A rich line of work has produced estimators of intrinsic dimension with complementary strengths.
\emph{TwoNN} estimates ID from ratios of first/second neighbor distances via a simple log–log fit \cite{Facco2017TwoNN};
\emph{MLE} fits a local Poisson model to $k$-NN distances;
\emph{TLE} leverages extreme-value theory in tight neighborhoods to reduce variance \cite{Amsaleg2022TLE}.
Beyond distance methods, \textsc{PHDim} uses persistent homology to capture multi-scale connectivity/holes and is comparatively stable under text domain shift and addition of noise to embeddings \cite{tulchinskii2023}.
The formal definitions are given in App.~\ref{app:id}.

\subsection{Relation between ID and model-prediction entropy}

Intrinsic dimension (ID) is computed from the geometry of hidden representations and thus captures text complexity \emph{from the model’s embedding space}, complementing prediction-based metrics such as (cross-)entropy \cite{viswanathan2025geometrytokensinternalrepresentations}. We show that ID and entropy are \emph{complementary}: ID encodes structural information not reflected in entropy.

\begin{figure*}[t!]
  \centering

  \begin{minipage}[t]{0.4\textwidth}
    \centering
    \includegraphics[width=\linewidth]{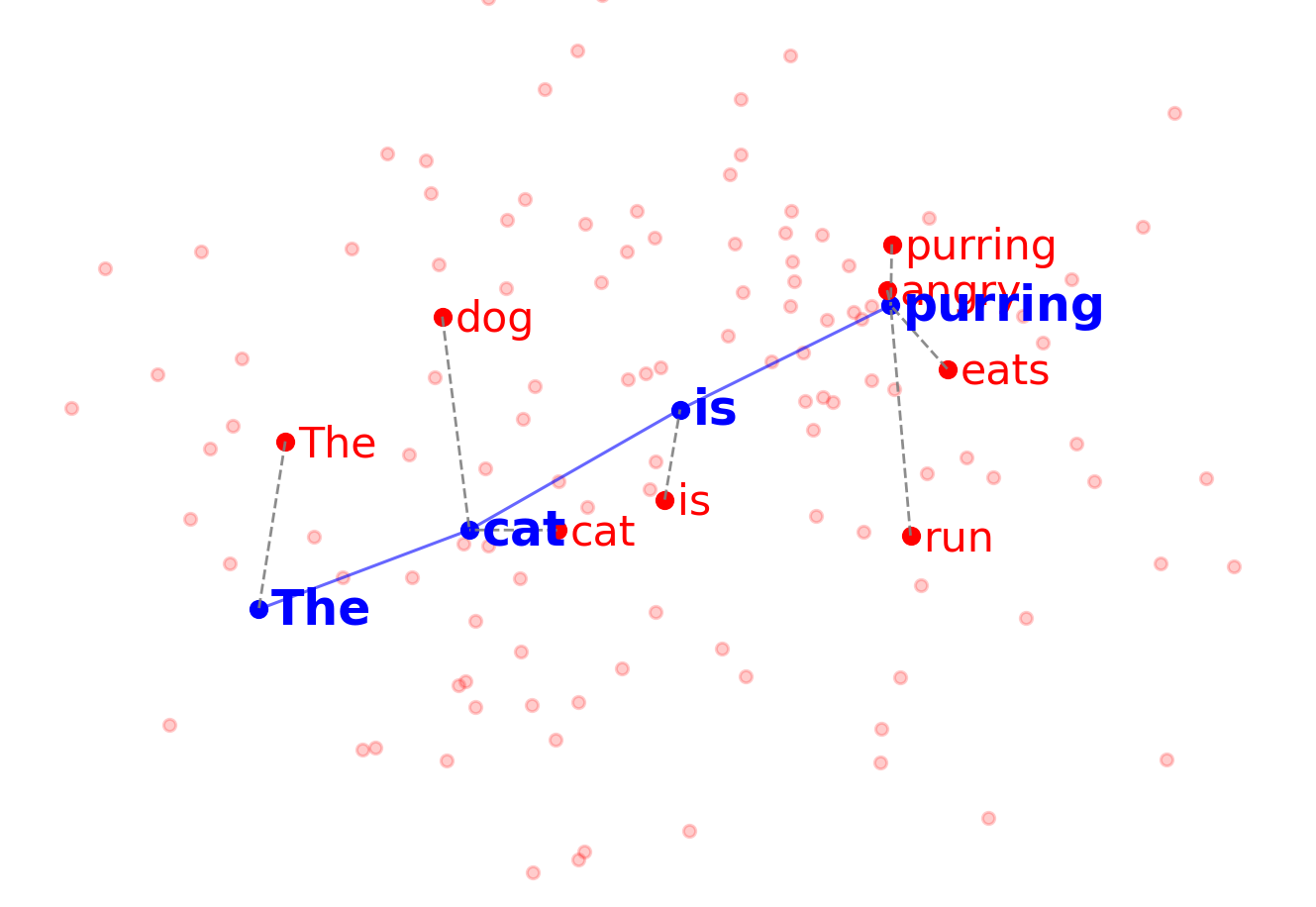}\\[2pt]
    
  \end{minipage}\hfill
  \begin{minipage}[t]{0.295\textwidth}
    \centering
    \includegraphics[width=\linewidth]{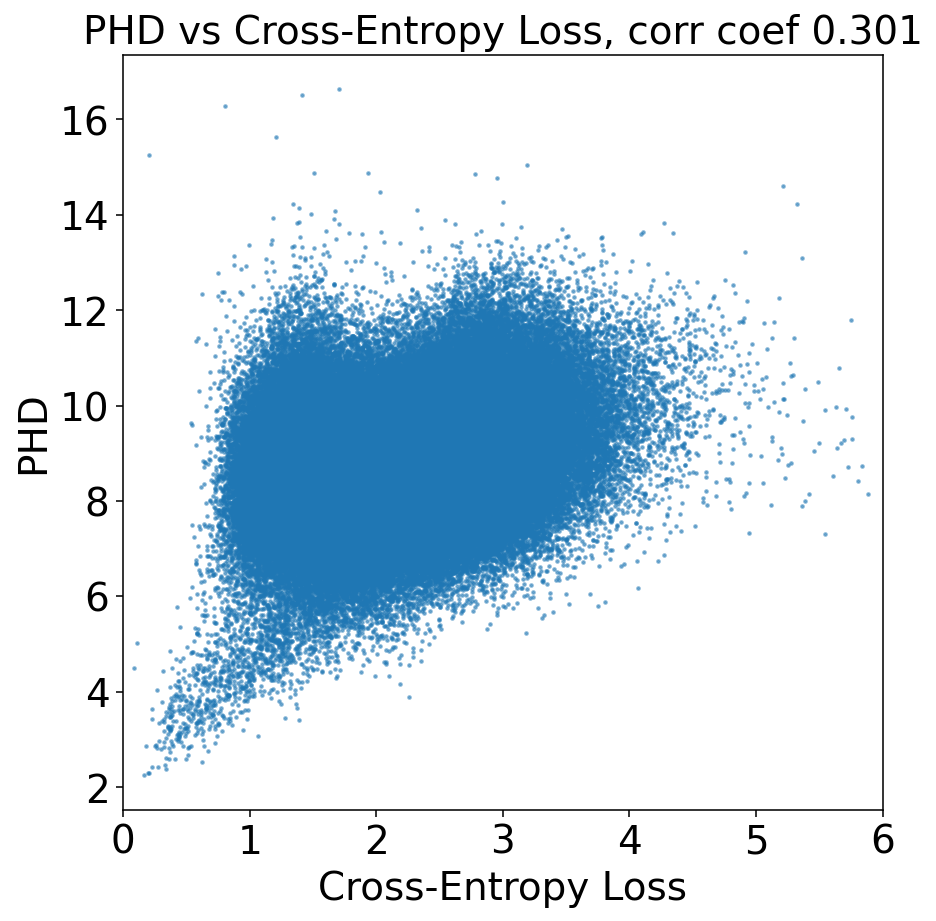}\\[2pt]
  \end{minipage}\hfill
  \begin{minipage}[t]{0.3\textwidth}
    \centering
    \includegraphics[width=\linewidth]{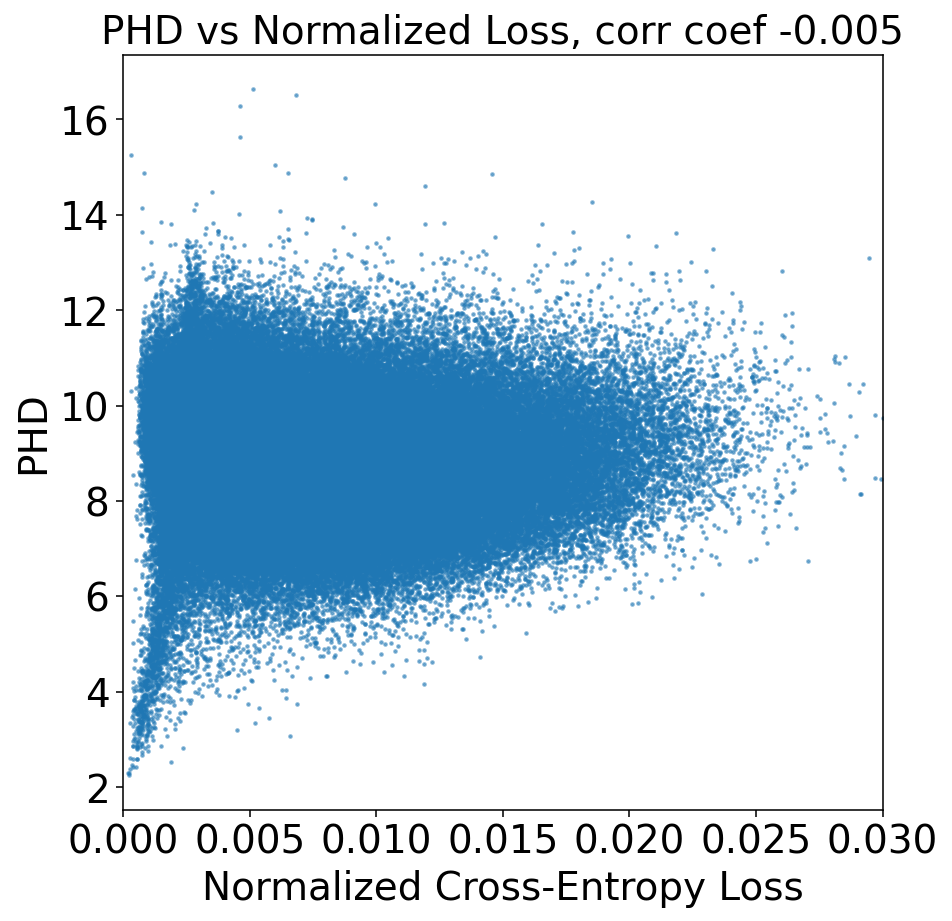}\\[2pt]
  \end{minipage}

  \caption{Intrinsic dimension characterizes the geometry of hidden representations (blue points on the leftmost frame), while prediction-based metrics such as entropy and cross-entropy depend on the unembedding dictionary (red points on the leftmost frame). 
    Sequences of embeddings with the same intrinsic dimension may yield very different prediction entropies, 
    depending on how densely the unembedding vectors populate the surrounding space 
    (i.e., on the number of close neighbors, shown by grey connections). Note significant correlation between PHD and Cross-Entropy loss (center frame) and weak correlation between PHD and Cross-Entropy loss, normalized by text length in gemma tokens (rightmost frame).}
  \label{fig:id_and_cross_entropy}
\end{figure*}

Formally, let $h_{1:T}$ be last-layer embeddings (the point cloud used for ID). Entropy arises only after \emph{unembedding}: logits $z_t = W^\top h_t + b$ and softmax. Hence, entropy depends on the \emph{alignment} of $h_t$ with the vocabulary matrix $W$, whereas ID depends solely on the geometry of $\{h_t\}$. Figure~\ref{fig:id_and_cross_entropy} (left) illustrates this consideration: identical geometric structure (same ID) can yield different entropies if the local density of unembedding vectors differs.

Consequently, ID cannot be replaced by entropy: synonym-rich neighborhoods raise entropy without changing geometry; conversely, structurally complex but predictable text can have low entropy yet high ID (see App.~\ref{app:entropy_vs_dim}). Empirically, after controlling for length, ID and entropy are essentially uncorrelated, indicating that they capture distinct facets of complexity (Fig.~\ref{fig:id_and_cross_entropy}, right); a residual dependence remains only in the very low-ID regime.

\section{Intrinsic Dimension through the Lens of Text Semantics}\label{sec:semantic_background}

We aim to characterize which semantic and stylistic properties of text are associated with higher or lower intrinsic dimension (ID). Our methodology combines sparse, interpretable representations learned from model activations with causal interventions and independent linguistic diagnostics.

\subsection{Sparse Autoencoders (SAEs) and feature steering}

SAEs learn sparse, approximately monosemantic features from LLM activations, addressing polysemantic neurons \cite{Olah2020ZoomIn} under the superposition view \cite{Elhage2022Superposition}. Given a layer activation $\xvec\in\mathbb{R}^d$, the encoder--decoder \cite{Sharkey2023SAE}
\begin{equation}
\fvec=\sigma(\Wenc\xvec+\benc),\quad \xh=\Wdec\fvec+\bdec
\end{equation}
produces a sparse, nonnegative code $\fvec$ over decoder columns $\dvec_j$ (feature directions), with $\xvec\approx \bdec+\sum_j \fvec_j \dvec_j$. For a text, we aggregate a feature’s tokenwise activations (e.g., sum) to obtain its sequence-level value.

To probe causality, we modify the model’s hidden state during generation using a chosen SAE feature \cite{kuznetsov-etal-2025-feature}:
\begin{equation}
\xvec'=\xvec+\lambda\,A_i\,\dvec_i,
\end{equation}
where $A_i$ is the feature’s reference-scale (max activation on a held-out set) and $\lambda$ controls intervention strength.

SAE features give interpretable axes of variation; steering tests their \emph{causal} effects on text style and content. We use this to examine features correlated with intrinsic dimension (ID) and to validate whether they systematically increase or decrease representational complexity in generated text.

\paragraph{SAE features and selection.}
Building on evidence that sparse autoencoders (SAEs) separate semantic differences between synthetic and natural texts~\cite{kuznetsov-etal-2025-feature}, we utilize SAEs trained on LLM activations to obtain sparse latent features. For each feature, we aggregate token-level activations to the sequence level and measure its correlation with ID. Features with the strongest positive and negative correlations are retained for analysis.

\paragraph{Interpreting and probing features.}
To ascribe semantic meaning to the selected features, we: (i)~perform \emph{extremal-activation} analysis by contrasting texts across activation quantiles; (ii)~assess \emph{domain specialization} via average activations over diverse human-written corpora; and (iii)~conduct \emph{steering} experiments, applying small interventions along a feature direction during generation to observe controlled changes while preserving fluency. The three perspectives jointly provide convergent evidence about how each feature relates to ID.

\paragraph{External linguistic validation.}
We relate ID to independent text properties using TAACO \cite{taaco_paper} -- the tool for analysing lexical diversity and cohesion, e.g., moving-window type–token ratios and sentence-to-sentence overlap (see Appendix~\ref{sec:id_taaco} for details) -- and syntactic metrics: syntactic diversity measured as average pairwise distance between dependency-graph representations via a Weisfeiler–Lehman kernel \cite{guo2025benchmarkinglinguisticdiversitylarge}, and the POS compression ratio based on gzip over POS-tag sequences \cite{shaib-etal-2024-detection} -- see Appendix~\ref{app:syntax_div}. These diagnostics corroborate SAE-based findings and ground ID variation in observable lexical and syntactic structure.

\section{Experimental details}\label{sec:details}

\textbf{Data.}
We use the GenAI Content Detection Task~1 dataset of \citet{wang-etal-2025-genai} (henceforth, \textsc{COLING}). 
Unless stated otherwise, experiments run on a cleaned split of the development set: from 261{,}758 texts (human and multiple LLM families), we retain 172{,}750 after filtering out (i) samples with $<150$ \textsc{Gemma} tokens and (ii) cases where any \textsc{PHDim} estimator on \textsc{Gemma-2-2B}, \textsc{Qwen-2.5-1.5B}, or \textsc{RoBERTa-base} was unstable or outside $[2,18]$. 
\textsc{COLING} spans diverse domains (news, Wikipedia articles, essays, QA, summaries, reviews, scientific abstracts, forums, technical documentation) and generations from T5/T0, GPT, LLaMA, OPT, Mixtral, and others. Most semantic interpretation experiments are performed on the subset of Human-written texts.

For proficiency analysis we use \textsc{OneStopEnglish} \cite{vajjala-lucic-2018-onestopenglish}: 189 human-authored texts, each in Elementary/Intermediate/Advanced versions (567 total).
For controlled generation we use \textsc{RAID} \cite{dugan-etal-2024-raid}: (i) 1{,}000 prompts with temperatures $0.2$--$2.0$ for temperature–ID studies; (ii) $20$ prompts from each of five domains (abstracts, books, news, Reddit, Wikipedia) for fine-grained steering.

\textbf{SAE setup.}
We employ \textsc{Gemma-Scope} \cite{lieberum-etal-2024-gemma} SAEs for \textsc{Gemma-2} \cite{gemmateam2024gemma2improvingopen}, using the canonical residual SAE (width $16$k) on \textsc{Gemma-2-2B} (base and instruct). 
For steering, we apply $\lambda\in\{0.25,0.5,1,5,10,15\}$ and report the largest $\lambda$ per feature that preserves coherence (very large $\lambda$ can induce repetitive or broken outputs).

\begin{figure}[t!]
        \centering
        \includegraphics[width=0.99\linewidth]{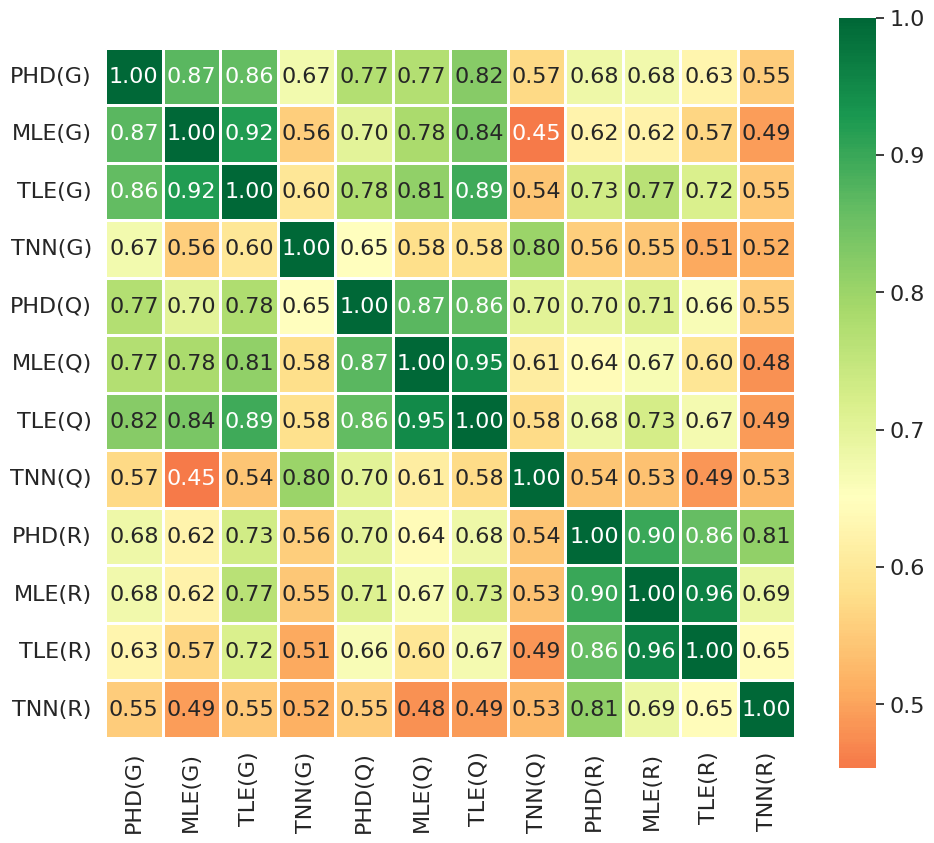}
        \caption{Correlations among various ID estimators. (G) denotes ID estimators upon Gemma, (R) - RoBERTa, (Q) - Qwen. %
        Note that PHD estimators upon all three models have correlation more than $0.5$ with all other estimators, making it a solid compromise. See Appendix~\ref{sec:id_scatterplots} for scatterplots and further discussion.}
        \label{fig:id_heatmaps}
\end{figure} 

\section{Results}\label{sec:results}

\subsection{Comparison of ID estimators}

We compare \textsc{PHDim}, \textsc{MLE}, \textsc{TLE}, and \textsc{TwoNN} on \textsc{Gemma}, \textsc{Qwen}, and \textsc{Roberta} embeddings (Fig.~\ref{fig:id_heatmaps}). All estimators are substantially correlated (pairwise $r>0.45$). \textsc{TwoNN} deviates most, whereas \textsc{PHDim} tracks the others best ($r>0.67$ with each on the same embeddings), so we use \textsc{PHDim} as the primary estimator.

Across embeddings, \textsc{Gemma} and \textsc{Qwen} agree closely for a fixed estimator, while \textsc{Roberta} diverges more; we revisit this in Appendix.

Within a model family, ID grows slightly with model size (Fig.~\ref{fig:phd_diff_sizes1}). Nevertheless, across methods, sizes, and families, well-formed texts concentrate in a narrow band (ID $\approx 5$--$15$), with outliers due to very short texts (Fig.~\ref{fig:scatter_plot}) or broken syntax. %
These IDs remain far below ambient embedding dimensions (hundreds to thousands).

\subsection{PHDim and data properties}

\textbf{Text length.}
We find \textsc{PHDim} to be stable for sufficiently long samples, with high variance only for short texts (App.~\ref{app:length}). Hence, we exclude texts shorter than 150 tokens.

\begin{figure}[t!]
        \centering
        \includegraphics[width=0.99\linewidth]{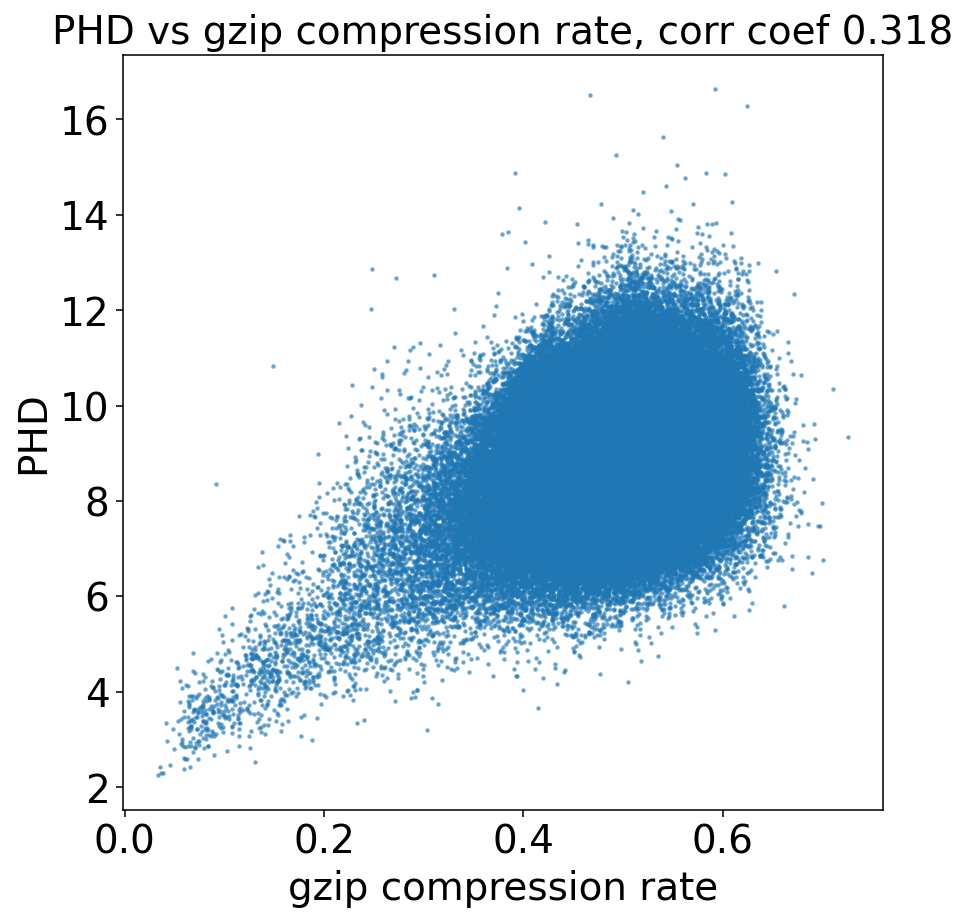}
        \caption{PHD and gzip}
        \label{fig:phd_vs_gzip}
\end{figure}

\textbf{Intrinsic data complexity.}
\textsc{PHDim} is positively correlated with gzip compression ratio (\(\mathrm{CR}\)); for texts longer than 150 tokens the correlation exceeds \(0.3\) (Fig.~\ref{fig:phd_vs_gzip}). The scatter exhibits a wedge-shaped support:
\(10\,\mathrm{CR}\ \lesssim\ \mathrm{ID}\ \lesssim\ 25\,\mathrm{CR}\).
Thus, \(\mathrm{CR}\) bounds but does not pointwise predict ID.

\textbf{Embedding geometry.}
Appendix~\ref{appnd:anisotropy} shows only weak correlations between ID and isotropy measures, indicating that radial/directional uniformity is not a driver of \textsc{PHDim}. By contrast, the correlation between \textsc{PHDim} and explained variance at rank $k$ (EV-$k$) is sharply peaked at $k\!\approx\!60$ (Fig.~\ref{fig:anisotropy_corr}). This suggests a \emph{global linear embedding dimension} of $\sim\!60$ despite a local ID of $\sim\!10$, highlighting strong manifold nonlinearity and cautioning against purely linear characterizations.

\begin{figure}[t!]
        \centering
        \includegraphics[width=0.99\linewidth]{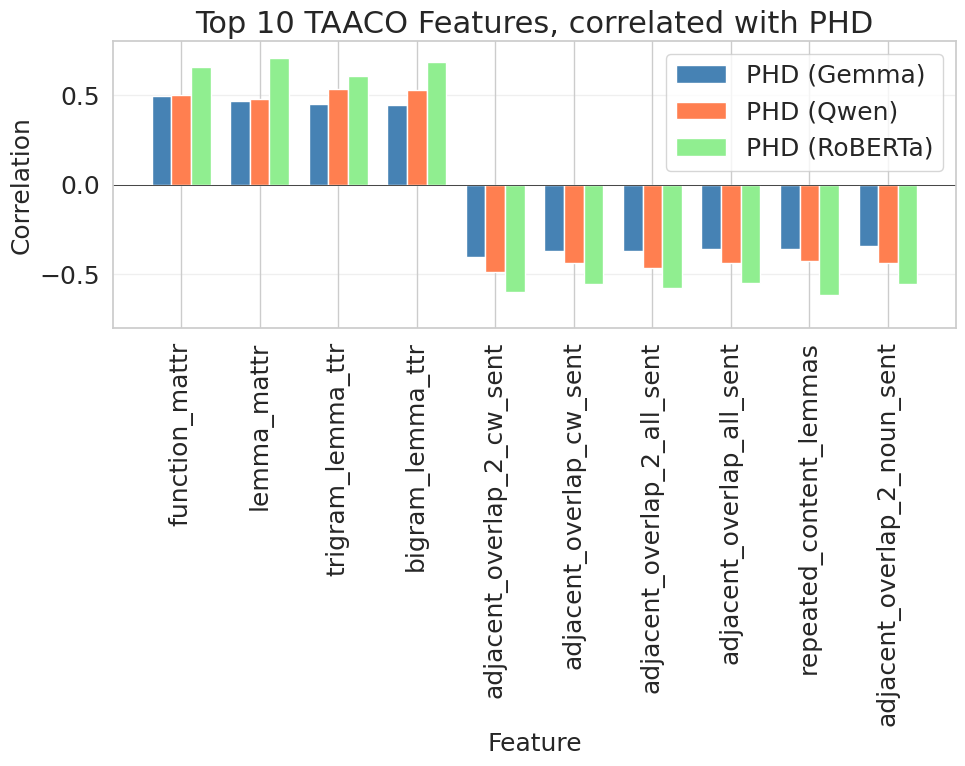}
        \caption{Top-10 features from TAACO with the strongest correlation with PHD(Gemma). See Appendix~\ref{sec:id_taaco} for similar barplots with MLE, TLE, TwoNN.}
        \label{fig:id_taaco_corr}
\end{figure}

\textbf{Lexical properties.}
\textsc{PHDim} correlates positively with lexical diversity (type--token ratio and $n$-gram diversity) and negatively with sentence-level overlap/repetition; syntax/discourse metrics show weak associations (Fig.~\ref{fig:id_taaco_corr}). Appendix experiments further show: (i) robustness to perturbations that preserve repetition patterns while destroying semantics; (ii) extreme out-of-range values on broken generations from weaker LLMs (very low for repetitive, very high for disconnected text); and (iii) limited informativeness of syntactic metrics for well-formed texts.

\subsection{ID divergence by text source and domain}

\begin{figure}[t!]
        \centering
        \includegraphics[width=0.99\linewidth]{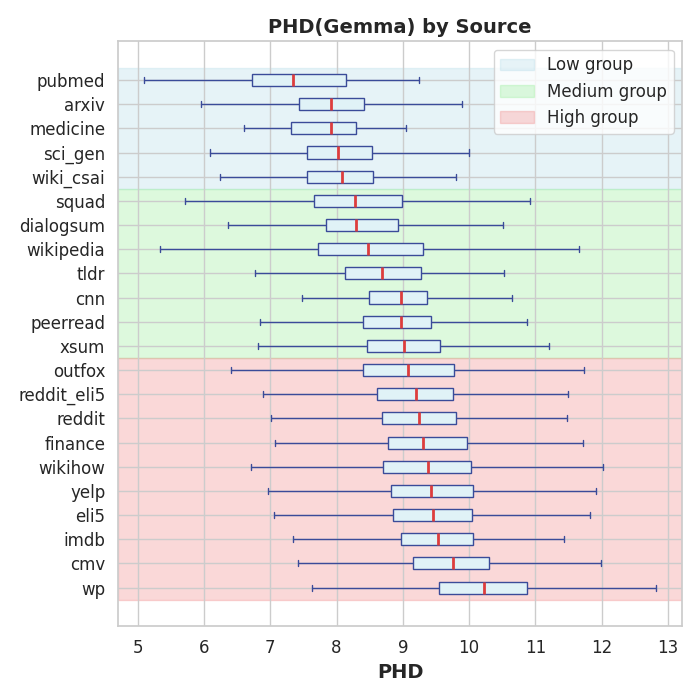}
        \caption{PHD(Gemma) by source with group differentiation.}
        \label{fig:phd_gemma_by_source}
\end{figure}

Contrary to~\citet{tulchinskii2023}, we observe a clear shift of ID values between domains (Fig.~\ref{fig:phd_gemma_by_source}).  The domains can be divided into three groups with distinct semantic and stylistic characteristics:

\begin{enumerate}
    \item \textbf{Low-dimensional group} (\emph{mean \textsc{PHDim} 7.5--8}): includes scientific and technical sources such as \texttt{arxiv}, \texttt{medicine}, \texttt{pubmed}, \texttt{wiki\_csai}, and \texttt{sci\_gen}.
    \item \textbf{Medium-dimensional group} (\emph{mean \textsc{PHDim} 8--9}): consists of factual and reporting texts with encyclopedic or formal news style, such as \texttt{cnn}, \texttt{xsum}, and \texttt{wikipedia}.
    \item \textbf{High-dimensional group} (\emph{mean \textsc{PHDim} 9--10.5}): includes essays, opinionated writing, and user-generated content with more informal or creative style, such as \texttt{cmv}, \texttt{eli5}, \texttt{wp}.%
\end{enumerate}

\begin{figure}[t!]
        \centering
        \includegraphics[width=0.99\linewidth]{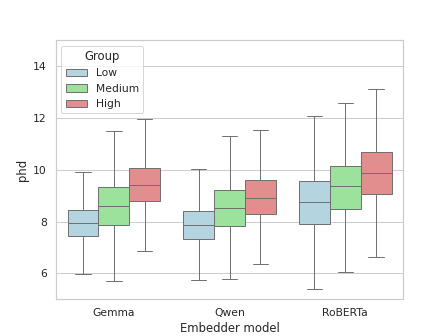}
        \caption{
            PHD by groups with different embedders models.
        }
        \label{fig:phd_groups_with_different_embedder_models}
\end{figure}

The lowest ID values appear in \texttt{pubmed} papers, while the highest (among public-domain corpora) occur in fictional stories from \texttt{wp}. Notably, \texttt{peerread} (OpenReview discussions), which blends scientific and opinionated styles, lies in the mid-\textsc{PHDim} range. The ordering of domains by \textsc{PHDim} is nearly identical across encoders (Fig.~\ref{fig:phd_groups_with_different_embedder_models}).

We next estimate \textsc{PHDim} for student-written stories at three proficiency levels—elementary, intermediate, and advanced (Fig.~\ref{fig:proficiency_phd}). \textsc{PHDim} rises with linguistic complexity: elementary texts have mean \textsc{PHDim} $9.5$ (near the lower edge of the \emph{fictional} group), whereas advanced texts approach or slightly exceed \texttt{wp} fiction (mean $12.5$). Although ID correlates with proficiency, genre is the stronger factor: even elementary \emph{fiction} surpasses \emph{scientific} and \emph{informational} genres.

These results imply that, for modern LLMs, scientific texts are the “simplest,” while informal, opinionated, and creative writing drives higher intrinsic dimensionality, peaking in fiction. This pattern aligns with our linguistic analysis: scientific prose shows strong topical/terminological coherence (high overlap/repetition, negatively correlated with \textsc{PHDim}), whereas fiction demands greater lexical diversity. Simplified syntax at the elementary level has little effect on embedding dimensionality.

\subsection{SAE-based interpretation}

\begin{figure*}[ht!]
        \centering
        \includegraphics[width=0.99\linewidth, ]{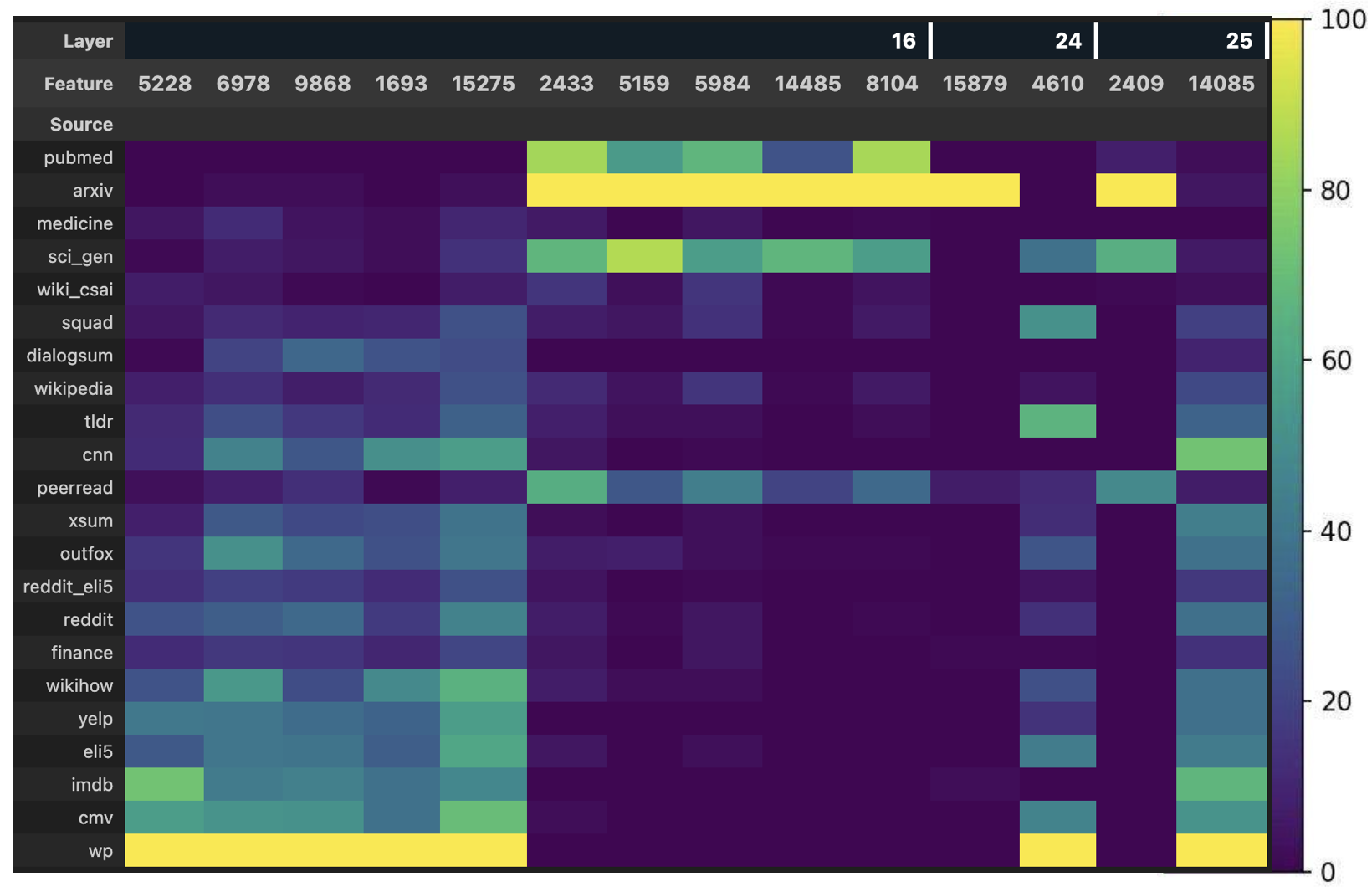}
        \caption{Colormap of SAE feature activations table by source of the text (domain) }
        \label{fig:sae_feature_colormap}
\end{figure*}

We selected a set of middle-layer features with the strongest absolute correlation with intrinsic dimension (ID), including both positively and negatively correlated cases. 
First, we computed mean activations across domains. Figure~\ref{fig:sae_feature_colormap} shows a clear bifurcation: a \emph{scientific} cluster (typified by \texttt{arXiv}) and a \emph{humanized} cluster with strongest activations in \texttt{wp} (writing prompts/short fiction). The former correlates negatively with ID, the latter positively, confirming a close link between ID and content (Table~\ref{tab:feature_interpretations}). We then interpreted features via (i) extremal-activation examples and (ii) targeted steering. The evidence is consistent: features \emph{positively} correlated with ID yield more emotional, personalized, rhetorically complex text, whereas \emph{negative} features push generations toward academic/reporting styles (formal tone, rigid structure).

Layer-wise, late layers (24–25) set global discourse templates, e.g., movie-plot, social-media article, BBC-style report, analytical template (cf.\ \texttt{14085}, \texttt{4610}, \texttt{15879}, \texttt{2409}), whereas layer~16 modulates local stylistic details (e.g., scientific framing \texttt{5159}, population statistics \texttt{8104}, claim foregrounding \texttt{2433}). Among informal signals, \texttt{5228} induces an uncertain or subdued tone, \texttt{9868} injects engaging elements into abstracts, and \texttt{6978} steers toward a “writing-assistant’’ voice. Table~\ref{tab:steering_examples} shows concise steering transformations; additional examples appear in App.~\ref{app:steering}.

\begin{table*}[ht!]
\footnotesize
\centering
\begin{tabular}{p{1.3cm}p{3cm}p{0.5cm}p{4.5cm}p{4.3cm}}
\toprule
\textbf{Feature} & \textbf{Domain} & r & \textbf{Extreme values} & \textbf{Steering} \\
\midrule

\multicolumn{5}{l}{\emph{Typical features with negative correlation with ID:}}\\[1mm]
16-5984 &  arxiv, sci-gen & $-0.3$ & Polished, formal, logically structured documents (e.g., \textcolor{blue}{academic}, reports) & \textcolor{blue}{Academic reframing}; nominalizations; enumerations \\
16-14485 & arxiv, sci-gen & $-0.36$ &  Complex arguments, structured \textcolor{blue}{academic} or expository texts & \textcolor{blue}{Formal academic tone}; expository structure; sectioning \\
16-2433 & arxiv, pubmed, sci-gen, peerread  & $-0.35$ & Unified compositions with \textcolor{blue}{clear rhetorical purpose} & Outline/placeholder scaffold; bulletization; meta-headers \\
16-8104 & arxiv, pubmed, sci-gen & $-0.33$ & Continuous \textcolor{blue}{formal prose} (abstracts, summaries) & \textcolor{blue}{Social-science} framing; \textcolor{blue}{survey stats}; institutional voice \\
16-5159 & arxiv, sci-gen & $-0.34$ &  \textcolor{blue}{Academic/scientific} style, technical vocabulary, objective tone & Analytical reportese; self-reference; repetitive passive \\
24-15879 & arxiv & $-0.31$ &  Highly \textcolor{blue}{formal}, technical, objective \textcolor{blue}{academic} prose & BBC/news article template; anecdotal lede; call-to-action \\
25-2409 & arxiv, sci-gen, peerread & $-0.43$  & \textcolor{blue}{Formal}, self-contained, expository prose & Stronger macro-structure; cohesive sections; ‘‘finally’’ cadence \\
\multicolumn{5}{l}{\emph{Typical features with positive correlation with ID:}}\\[1mm]
16-15275 & wp, cmv, wikihow, eli5, yelp, cnn & $0.36$  &  Coherent, human-readable, publication-quality texts & \textcolor{red}{Personalization}; concrete characters; richer plot detail \\
16-5228 & wp, imdb & $0.33$ &  \textcolor{red}{Conversational}, emphatic, lengthened texts with repetition & \textcolor{red}{Uncertainty/hedging; introspection}; open-endedness \\
16-6978 & wp, wikihow, outfox, cnn & $0.36$ & \textcolor{red}{Multi-source}, composite, sometimes incoherent synthesis & \textcolor{red}{Informal/meta voice}; bracketed notes; assistant aside \\
16-9868 & wp & $0.37$ & \textcolor{red}{Informal}, noisy, user-generated style; robust to corrupted input & \textcolor{red}{Media/teaser tone}; punchy fragments; promo cadence \\
16-1693 & wp, cnn, wikihow & $0.37$  & Long, elaborated, \textcolor{red}{multi-threaded} arguments/narratives & \textcolor{red}{Forum/discussion vibe}; colloquial; community framing \\
25-14085 & wp, cnn, imdb & $0.4$ & \textcolor{red}{Personal voice}; 1st/2nd person; informal, conversational, opinionated; narratives and reviews. & \textcolor{red}{Movie/outline format}; headings; character bios \\
24-4610 & wp, tldr &  $0.36$ & Paragraphs or list items start with \textcolor{purple}{`,`, `,,`}, sometimes exaggerated \textcolor{purple}{`,,,,`} & \textcolor{red}{Social-media article}; bold subheads; quotes + how-to steps \\

\bottomrule
\end{tabular}
\caption{Short interpretations of selected features and their effect on document style depending on activation.}
\label{tab:feature_interpretations}
\end{table*}

\begin{table*}[h!]
\footnotesize
\begin{tabular}{p{1.3cm}p{0.3cm}p{3.5cm}p{9.2cm}}
\hline
\textbf{Feature} & \textbf{Sign} & \textbf{Shift} & \textbf{Example change (steered)} \\
\hline
\multicolumn{4}{p{15.3cm}}{
\textbf{Default generation (excerpt):} 

\emph{The classic Ford Falcon, a symbol of American muscle in the 1960s, is a popular choice for car enthusiasts. \dots{}
The problem can stem from a number of factors, including a clogged fuel line, a faulty fuel pump, or a blocked fuel filter.}

}\\[1mm]
16-14485 & -- & Research-report structure & ``The article identifies three primary causes for the fuel supply deficiency; firstly..., secondly..., thirdly....'' \\
16-5159 & -- & Analytical, self-referential & ``This paper examines potential causes and provides a diagnostic framework; this paper is intended to be a guide.'' \\
16-15275 & + & Storytelling, humanized & ``John, a 30-year restoration veteran, replaced the pump, lines, and filter — yet the Falcon still refuses to start.'' \\
16-1693 & + & Forum-like conversational & ``This can be a real headache — we’re gonna get that carb working again!'' \\
\hline
\end{tabular}
\caption{Steering effect of several representative features. ``Sign'' reflects if the feature positively or negatively correlated with intrinsic dimension}\label{tab:steering_examples}
\end{table*}

\section{Discussion}\label{sec:discussion}

Intrinsic dimension (ID) remains an underexplored measure of textual complexity. Our extensive experiments (with details in the Appendix) elucidate its semantic and structural correlates and support the following conclusions.

\textbf{Complementarity to prediction-based metrics.}
ID is complementary to (cross-)entropy and related prediction-based measures: after controlling for length, the two are largely uncorrelated, indicating that ID captures geometric/structural information not reflected in likelihood.

\textbf{Narrow operating band, far below ambient dimension.}
For well-formed texts, ID concentrates in a narrow range ($\approx$5–15) and is remarkably stable across domains, encoder families/sizes, and ID estimators. Yet ID \emph{cannot} be substituted by linear dimensionality reduction: the manifold is highly curved, with an effective \emph{global linear embedding dimension} of $\sim$60–100 (e.g., peak PHDim–EV$_k$ correlation near $k\!\approx\!60$).

\textbf{Failure detection.}
ID reliably flags broken generations: extremely low values for repetitive/looped text and extremely high values for disconnected fragments or incoherent mixtures.

\textbf{Compression is bounding, not predictive.}
ID correlates moderately with gzip compression ratio, but the relationship is not pointwise predictive; compression provides bounds on plausible ID rather than accurate estimates.

\textbf{Linguistic correlates.}
ID increases with lexical diversity and decreases with cross-sentence repetition/overlap; it is comparatively insensitive to syntactic indicators for well-formed text.

\textbf{Style and genre effects.}
ID is strongly conditioned by style/genre. Scientific and purely informational prose exhibits lower ID, whereas personalized, opinionated, and fictional writing attains the highest ID. Adding “semantic dimensions’’, such as personality, emotion, narrative/plot, stance, informality, raises ID.

\paragraph{Implications and cautions.}
Prior work has used ID for training dynamics evaluation, synthetic text detection, scaling-law validation, architecture search, and memorization studies. However, treating ID as a monolithic proxy for “difficulty’’ risks misinterpretation: domain shifts, spurious correlations, and evaluations confined to low-ID corpora (e.g., Wikipedia-like text) can yield misleading conclusions. Our findings indicate that complexity-based LLM analyses should explicitly target higher-ID domains (forums, fiction, opinionated writing) and report local (ID) linear structure, rather than relying solely on linear projections or prediction-based metrics.

\section{Conclusion}
In this work, we have analyzed intrinsic dimensionality as a lens through which to interrogate the geometric structure of text representations in large language models. Our findings reveal that ID is not a monolithic indicator of complexity, but rather a domain- and style-sensitive property: scientific and informational texts occupy low-ID regions, while opinionated, narrative, and personalized writing exhibit substantially higher ID, reflecting greater semantic degrees of freedom. Critically, ID is largely orthogonal to prediction-based measures like cross-entropy, underscoring its value in exposing structural regularities that likelihood alone cannot capture.

\section{Limitations}
 Our PHD analysis uses embeddings from only three models (Gemma, RoBERTa, Qwen); results may vary with other encoders, as intrinsic dimension estimates are sensitive to embedding choice. Moreover, PHD is computed on random subsamples of the data, and due to this stochastic sampling procedure, PHD estimates may exhibit some variability across runs.
 Finally, common estimators—TLE, MLE, TwoNN, and PHD—rest on different assumptions about the geometry and distribution of token embeddings. Consequently, these estimators capture complementary but non-equivalent aspects of intrinsic dimensionality and are not directly comparable.

\bibliography{custom}

\appendix

\section{Mathematics behind Intrinsic Dimension}

\subsection{Intrinsic dimension definitions}\label{app:id}

We outline several common notions of intrinsic dimension used in data analysis, and illustrate them with uniformly distributed points in a $d$-dimensional Euclidean space. Throughout, $\sim$ denotes asymptotic proportionality.

\paragraph{MLE (Maximum Likelihood Estimation dimension).}  
For a point $x$ consider the probability that a random point $x'$ falls in its $\varepsilon$-neighborhood. For a uniform distribution, this probability scales as:
\begin{equation}
    \Pr(\|x-x'\|_2 < \varepsilon) \;\sim\; C(x)\,\varepsilon^d. \tag{1}
\end{equation}

Hence, the exponent in~Eq.~(1) defines the \emph{MLE dimension}. Formally, we compute
\begin{equation}
\dim_{\mathrm{MLE}} = 
\mathbb{E}_{x}\left[\lim_{\varepsilon \to 0}
\frac{\log N_\varepsilon(x)}{\log 1/\varepsilon}\right], \tag{1a}
\end{equation}
where $N_\varepsilon(x)$ is the number of points in the $\varepsilon$-neighborhood of $x$.  

\paragraph{TwoNN (Two Nearest Neighbors dimension).}  
For each point $x$, let $r_1$ and $r_2$ be the distances to its nearest and second-nearest neighbors, and define the ratio  
\[
    \mu = r_2/r_1.
\]  

For uniform distribution in $\mathbb{R}^d$, the following holds:  
\begin{equation}
\Pr(\mu \le t) = 1 - t^{-d}, \quad t \ge 1. \tag{2}
\end{equation}  
Thus, the distribution of $\mu$ depends only on $d$.
We can estimate $d$ by fitting this model CDF to the empirical CDF $\hat{F}(\mu)$, for example by minimizing a divergence:
\begin{align}
\dim_{\mathrm{TwoNN}} & = \nonumber\\ 
\arg&\!\min_{d} \; D_{\mathrm{KL}}\!\left(\hat{F}(\mu) \,\|\, (1 - \mu^{-d})\right), \tag{2a}
\end{align}
where $\hat{F}(\mu)$ is the empirical cumulative distribution function of the ratios.  

\paragraph{PHDim (Persistent Homology Dimension).}  
Let $N_\varepsilon$ be the number of nontrivial homological features (e.g., connected components or cycles) that appear in a simplicial complex built from the data at scale $\varepsilon$. For a set in $\mathbb{R}^d$ we get 
\begin{equation}
    N_\varepsilon \;\sim\; C\,\varepsilon^{-\dim_{\mathrm{PH}}}. \tag{3}
\end{equation}  
Thus, the \emph{PH dimension} is defined as
\begin{equation}
\dim_{\mathrm{PH}} = \limsup_{\varepsilon \to 0}
\frac{\log N_\varepsilon}{\log 1/\varepsilon}. \tag{3a}
\end{equation}

An equivalent characterization uses the total length of the minimum spanning tree (MST). Let $L_n$ be the MST length on $n$ points sampled uniformly from the set.
Asymptotically, one has
\begin{equation}
L_n \;\sim\; n^{\frac{d-1}{d}}, \tag{3b}
\end{equation}
which yields
\begin{equation}
\dim_{\mathrm{PH}} = \lim_{n \to \infty}
\frac{1}{1 - \frac{\log L_n}{\log n}}. \tag{3c}
\end{equation}

\subsection{Three Arguments Against Functional Dependence Between Entropy and Intrinsic Dimension}\label{app:entropy_vs_dim}

In prior work \cite{viswanathan2025geometrytokensinternalrepresentations}, 
the authors present several mathematical arguments suggesting a possible dependency 
between intrinsic dimension and entropy, although no formal proof is provided. 
In contrast, we offer three complementary arguments in the opposite direction. 
Each of them serves as a constructive counterexample, demonstrating that no 
functional dependency between entropy and intrinsic dimensionality can exist 
in the general case. 

Of course, neural networks do not represent the fully general case: 
their embeddings and parameter spaces are restricted by training dynamics 
and architectural constraints. 
However, our empirical results support these theoretical considerations, 
indicating that even in practical settings the observed correlations 
arise empirically rather than by necessity. 

\paragraph{(1) Dependence Argument.}
Let $U = \{u_i\}_{i=1}^V \subset \mathbb{R}^m$ be the unembedding matrix and 
$p(h) = \mathrm{softmax}(Uh)$ the model's prediction for a hidden state $h$.
The prediction entropy is
\[
H(h) = -\sum_i p_i(h)\log p_i(h),
\]
which depends explicitly on $U$. 

In contrast, the intrinsic dimension (ID) of a set $H \subset \mathbb{R}^m$ 
depends only on the \emph{geometry of embeddings}---it is a property of their 
relative arrangement in $\mathbb{R}^m$ and does not depend on the unembedding matrix $U$.

Even within a single model, one can select disjoint subsets of tokens 
whose projections under $U$ differ strongly, leading to distinct entropy distributions 
while the geometry of their embeddings (and thus ID) remains unchanged.

\smallskip
\noindent
\textbf{Example.} 
Assume that $U$ consists of two parts: 
a dense cluster of ``positive'' tokens $u_i$ with nonnegative components, 
and a single ``negative'' token $u_0 = (-1, -1, \dots, -1)$.
Consider a set $H_+$ of hidden vectors $h$ with positive coordinates.
For such $h$, all scalar products $u_i^\top h$ are positive, 
so $\mathrm{softmax}(Uh)$ is approximately uniform, and 
the average entropy $H(h)$ is close to $\log V$.
If we flip the signs of all these vectors, obtaining $H_- = \{-h : h \in H_+\}$,
the geometric structure and intrinsic dimension are identical,
but now all scalar products $u_i^\top (-h)$ are negative, 
so $\mathrm{softmax}(Uh)$ becomes nearly one-hot and $H(h)\approx 0$.
Thus, entropy can vary drastically while intrinsic dimension remains the same.

\paragraph{(2) Scale Invariance Argument.}
Consider a scaling map $T_{\alpha}(h) = \alpha h$. 
Since affine transformations preserve local dimensionality,
\[
\mathrm{ID}(T_{\alpha}(H)) = \mathrm{ID}(H).
\]
However, the corresponding prediction distribution transforms as
\[
p(T_{\alpha}(h)) = \mathrm{softmax}(U(\alpha h)) = \mathrm{softmax}(\alpha Uh),
\]
which is equivalent to inverse-temperature rescaling:
\[
\alpha \to 0 \Rightarrow p \to \tfrac{1}{V}\mathbf{1},
\qquad
\alpha \to \infty \Rightarrow p \to e_k,
\]
and therefore
\[
H(T_{\alpha}(h)) \in [0, \log V].
\]
Thus, entropy can be continuously adjusted from its maximum to minimum 
while intrinsic dimension remains invariant.
Entropy and ID are therefore \emph{affinely independent quantities}.

\paragraph{(3) Continuity Argument.}
With a fixed unembedding matrix, (token-level) entropy induces a continuous function
\(H:\mathbb{R}^m\!\to\!\mathbb{R}\) on the embedding space: it depends only on a point’s
location in the ambient space (hence on distances to the unembedding vectors), not on
its neighbors along the data manifold.

For a measurable set \(A\subset\mathbb{R}^m\), write
\(\bar H(A)\coloneqq |A|^{-1}\!\int_A H(h)\,dh\).
Let \(\mathcal M_d\) be a compact \(d\)-dimensional submanifold and
\(N_\varepsilon(\mathcal M_d)=\{x:\mathrm{dist}(x,\mathcal M_d)\le\varepsilon\}\) its tubular
\(\varepsilon\)-neighborhood, then we have:
\[
\bar H\!\big(N_\varepsilon(\mathcal M_d)\big)\xrightarrow[\varepsilon\to 0]{}\bar H(\mathcal M_d).
\]

Assume further that for any compact \(d\)-manifold \(\mathcal M_d\) the mean
\(\bar H(\mathcal M_d)\) depends only on \(d\) (denote this value by \(\mu(d)\)).
A compact \(D\)-manifold \(\mathcal M_D\) admit a finite cover by such tubular sets.
Then
\[
\bar H(\mathcal M_D)\;=\;\sum_{i=1}^N \alpha_i(\varepsilon)\,\bar H\!\big(N_{\varepsilon_i}(\mathcal M_d^{(i)})\big)
\;=\;\mu(d),
\]
where \(\alpha_i(\varepsilon)\!=\!|N_{\varepsilon_i}(\mathcal M_d^{(i)})|/\big|\!\bigcup_j N_{\varepsilon_j}(\mathcal M_d^{(j)})\!\big|\)
are convex weights.

\emph{Conclusion.} The mean entropy is continuous under infinitesimal “thickenings’’ of
dimension: it cannot exhibit jumps across intrinsic dimensions and therefore cannot be a
functional of intrinsic dimension alone.

\section{The connection between various ID Estimators}
\label{sec:id_scatterplots}

\subsection{MLE, TLE and TwoNN upon embeddings from main models}

\begin{figure*}[t]
        \centering
        \includegraphics[width=0.99\linewidth]{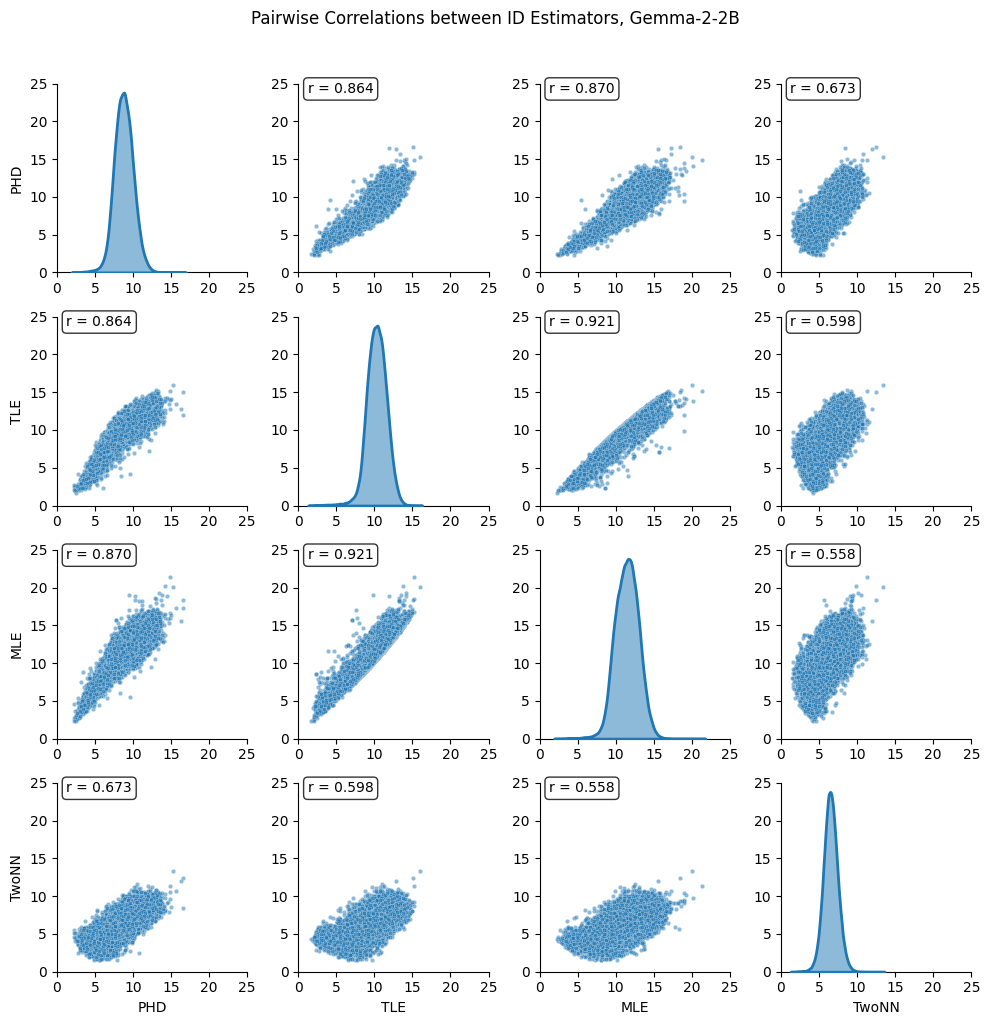}
        \caption{ID Estimators Gemma}
        \label{fig:id_estimators_gemma}
\end{figure*}

\begin{figure*}[t]
        \centering
        \includegraphics[width=0.99\linewidth]{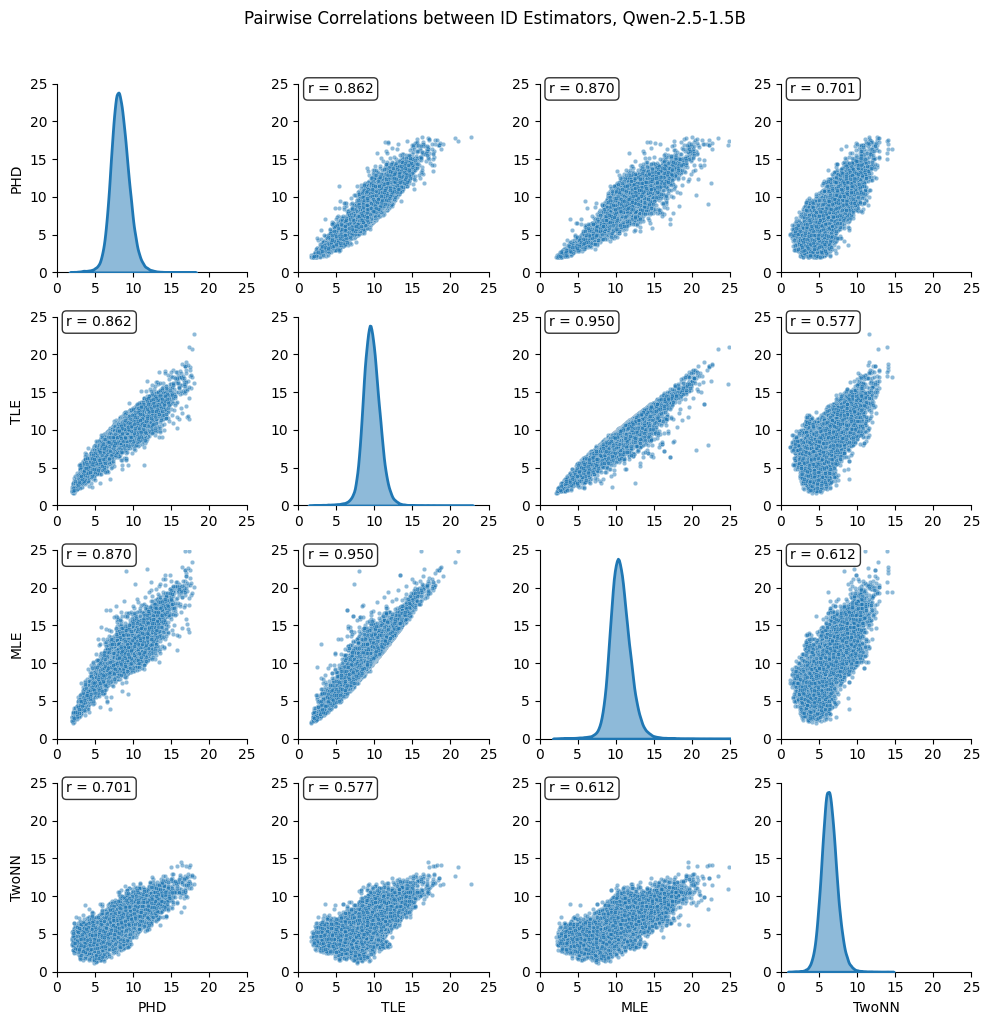}
        \caption{ID Estimators Qwen}
        \label{fig:id_estimators_qwen}
\end{figure*}

\begin{figure*}[t]
        \centering
        \includegraphics[width=0.99\linewidth]{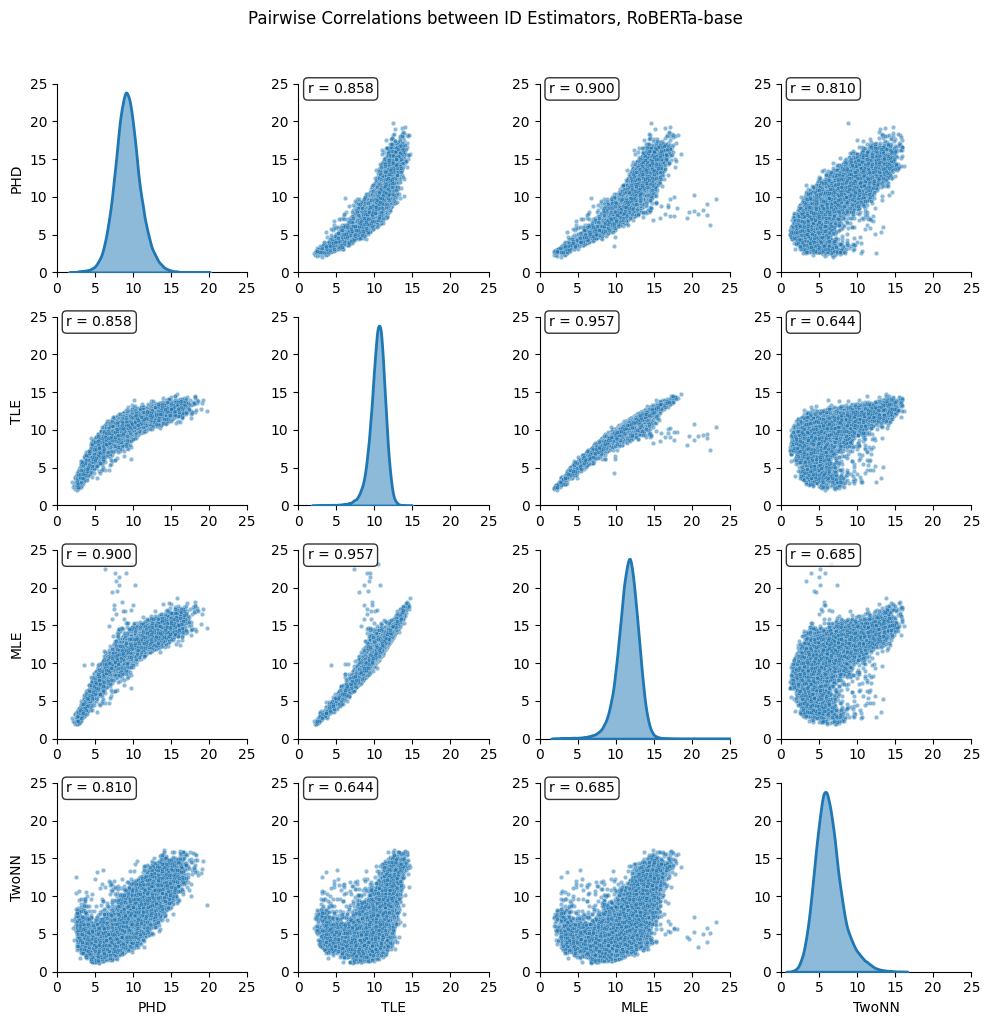}
        \caption{ID Estimators RoBERTa}
        \label{fig:id_estimators_roberta}
\end{figure*}

Figs.~\ref{fig:id_estimators_gemma}, \ref{fig:id_estimators_qwen}, and \ref{fig:id_estimators_roberta} illustrate pairwise relationships among intrinsic-dimension estimators computed on embeddings produced by the same model. MLE and TLE exibit very similar tracks but differ substantially from TwoNN. Notable, that PHD remains strongly correlated with all other estimators.

Most ID estimators are monotonically related; for Qwen and Gemma embeddings, the relationships are nearly linear.
RoBERTa stands out as an exception having substantially non‑linear relationships. The TwoNN estimator on RoBERTa is non‑monotonic with respect to the others.
See Fig.~\ref{fig:id_estimators_roberta} for details.

\subsection{PHD dependence on  model size}\label{app:phd_on_model_size}

We wondered how the number of parameters in the embedder model affects the internal dimension of the text. To ensure consistency, we used only Qwen 3 base models with sizes 0.6B, 1.7B, 4B, 8B, 14B, 32B. We calculated PHDim on human texts with a length of more than 150 tokens and obtained the following results, see Tab. ~\ref{tab:phd_diff_sizes1}.

We see a general trend towards an increase in PHDim as the model size increases. Qwen3-4B stands out from this picture.  This may be due to the fact that only this model has a 128k context window and uses the Embeddings Tie. For smaller models, the context window is 32k, and Embedding Tie is used. The Embedding Tie is no longer used for models, but the context window is 128k.\cite{yang2025qwen3technicalreport}.

\begin{table}[ht]
    \centering
    \begin{tabular}{lrrr}
    \toprule
    Model &  PHD & Emb. & Num.  \\
          & (median) & size & layers \\
    \midrule
    Qwen3-0.6B & 9.429 & 1024 & 28 \\
    Qwen3-1.7B & 9.534 & 2048 & 28 \\
    Qwen3-4B & 10.063  & 2560 & 36 \\
    Qwen3-8B & 9.601   & 4096 & 36\\
    Qwen3-14B & 10.046 & 5120 & 40 \\
    Qwen3-32B & 10.449 & 5120 & 64 \\
    \bottomrule
    \end{tabular}
    \caption{Median PHD dependence on Qwen3 size}
    \label{tab:phd_diff_sizes1}
\end{table}

\begin{figure}[H]
        \centering
        \includegraphics[width=\linewidth]{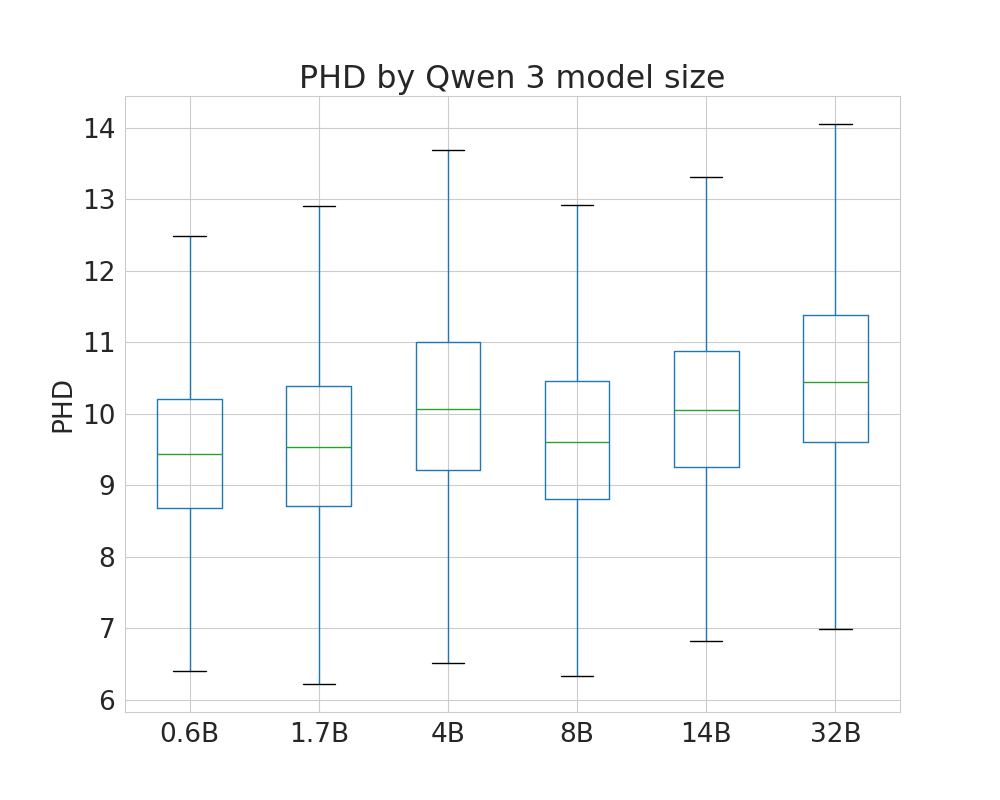}
        \caption{Boxplots of PHD across Qwen3 model sizes}
        \label{fig:phd_diff_sizes1}
\end{figure}

\subsection{Intrinsic Dimensionality and data geometry} %
\label{appnd:anisotropy} 

In this section, we study the connection between ID and other geometric properties of data in the embedding space, related to dataset complexity. 

There are several methods to measure the ``diversity'' (anisotropy) of a given dataset and here we will study their correlation with the intrinsic dimensionality.
Suppose we have a text and the set of its tokens embeddings $X = (x_1, \ldots , x_N), x_i \in \mathrm{R}^d$. As $\sigma_1, \sigma_2, \ldots, \sigma_r$ ($r = \min(N, d)$) we denote the singular values of the matrix of stacked embeddings $X$.

Following \cite{yusupov2025internalrepresentationstextquality}, we explore the following metrics:
\begin{itemize}
    \item \textbf{Maximal Explained Variance (MEV)}~\cite{razzhigaev-etal-2024-shape} -- the proportion of variance explained by the first principal component of the data (i.e., $\frac{\sigma_1^2}{\sum_{i=1}^r\sigma_i^2}$. Additionally, we calculate the proportion of variance, explored by the first $20$ components ($\frac{\sigma_1^2 + \ldots \sigma_{20}^2}{\sum_{i=1}^r\sigma_i^2}$, we denote it as \textbf{20-EV}); we properly explore the effect of the number of components further down in this section.    
    \item \textbf{Resultant Length}~\cite{ethayarajh-2019-contextual}. Let $x_i^\prime$ be the unit-normalized token embeddings (i.e., $x_i^\prime = \frac{x_i}{\lvert\rvert x_i\lvert\rvert_2}$, $i = 1..n$). Then, the Resultant Length of $X$ is the length of the mean directional vector 
    $
    R(X) = \lvert\rvert{\frac{1}{N}}\sum_{i=1}^N x_i^\prime\lvert\rvert_2
    $. For any text, $0 \leq R(X) \leq 1$, where $R = 0$ implies perfect isotropy and $R = 1$ implies that all tokens are perfectly aligned and lay on the same line.
    \item \textbf{Schatten-\textit{p} Norm}~\cite{bhatia1997matrix} quantifies the global spectral energy in the matrix $X$. It is defined as $\lvert\rvert X\lvert\rvert_{S_p} = \left(\sum_{i=1}^r\sigma_i^p\right)^{1/p} $  . In this section we study only the case of $p = 2$.
    \item \textbf{Effective Rank}~\cite{oliver2007erank} estimates the \textit{effective dimensionality} of the embedding space. It is proposed as an entropy-based continuous approximation of matrix rank that is robust to minor perturbations in the data.
    Let $p_k = \frac{\sigma_k}{\sum_{i = 1}^r\sigma_i}$ then Effective Rank of $X$ is calculated as $ERank(X) = \exp\left(-\sum_{i=1}^r p_k\log p_k\right)$
\end{itemize}

\begin{figure*}
    \centering
    \begin{tabular}{cc}
    \includegraphics[width=0.48\linewidth]{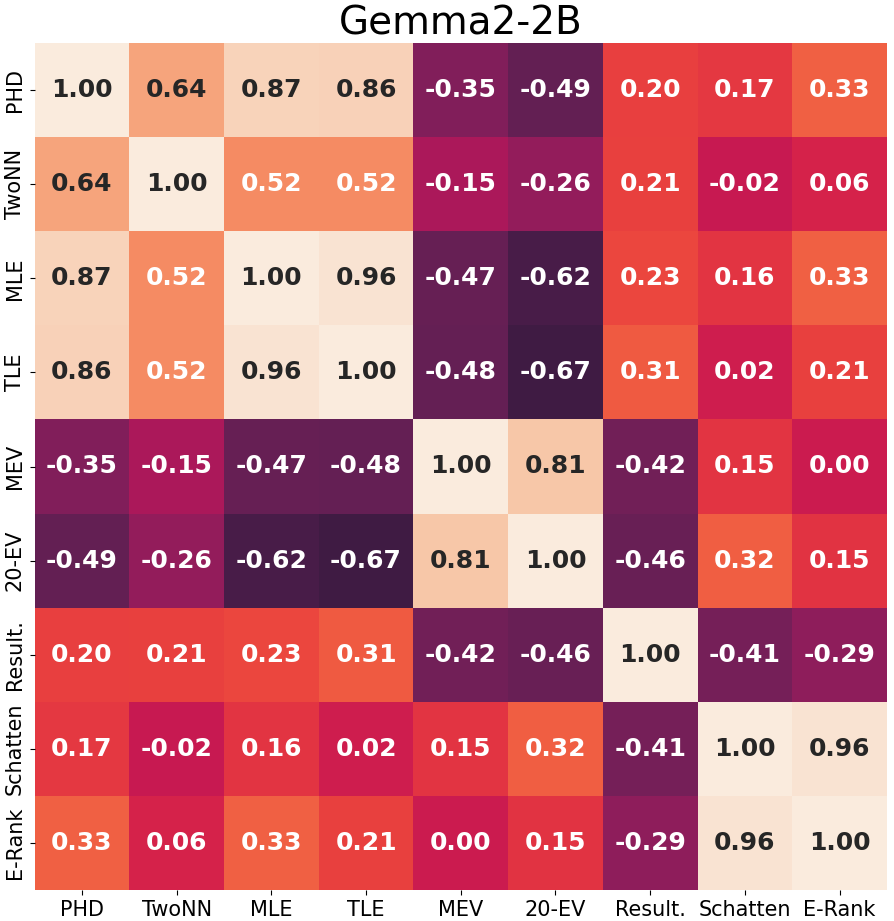} & 
    \includegraphics[width=0.48\linewidth]{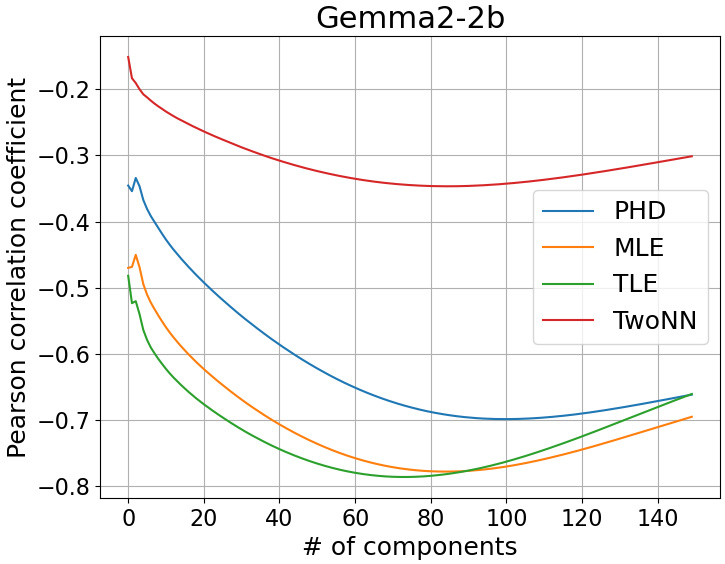}\\
    \includegraphics[width=0.48\linewidth]{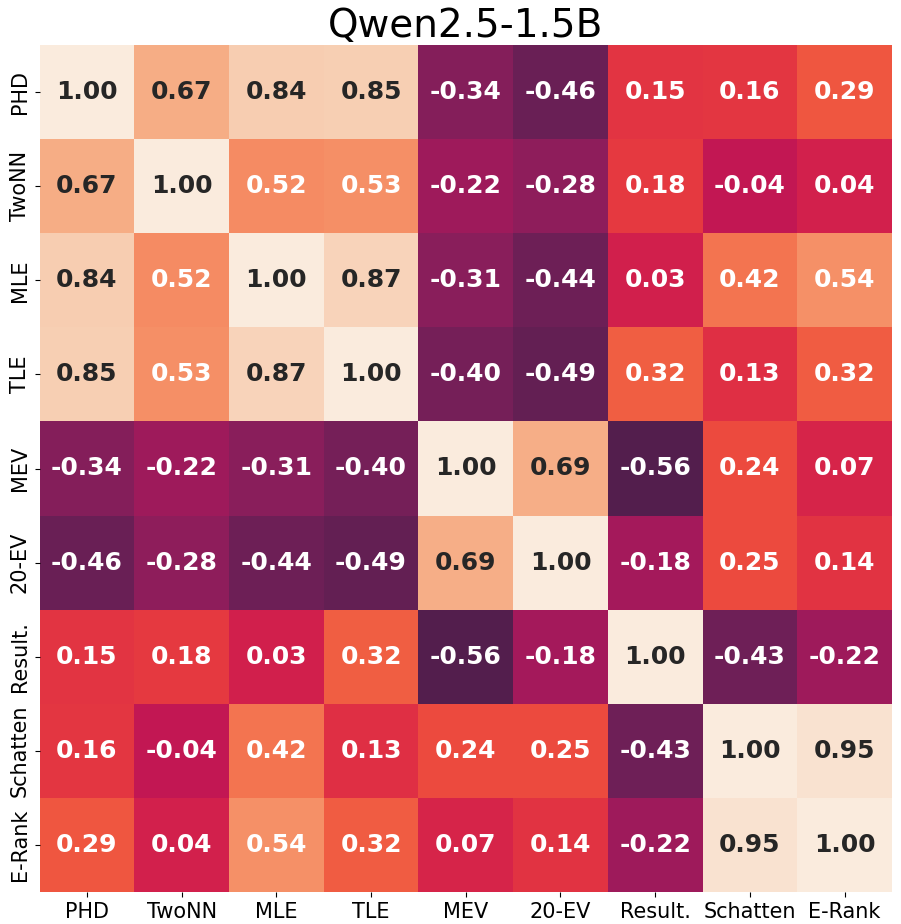} & 
    \includegraphics[width=0.48\linewidth]{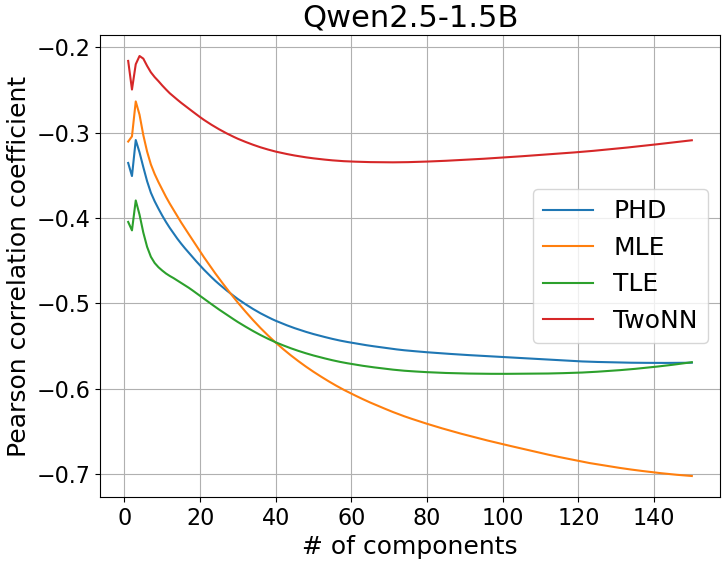} \\
    \includegraphics[width=0.48\linewidth]{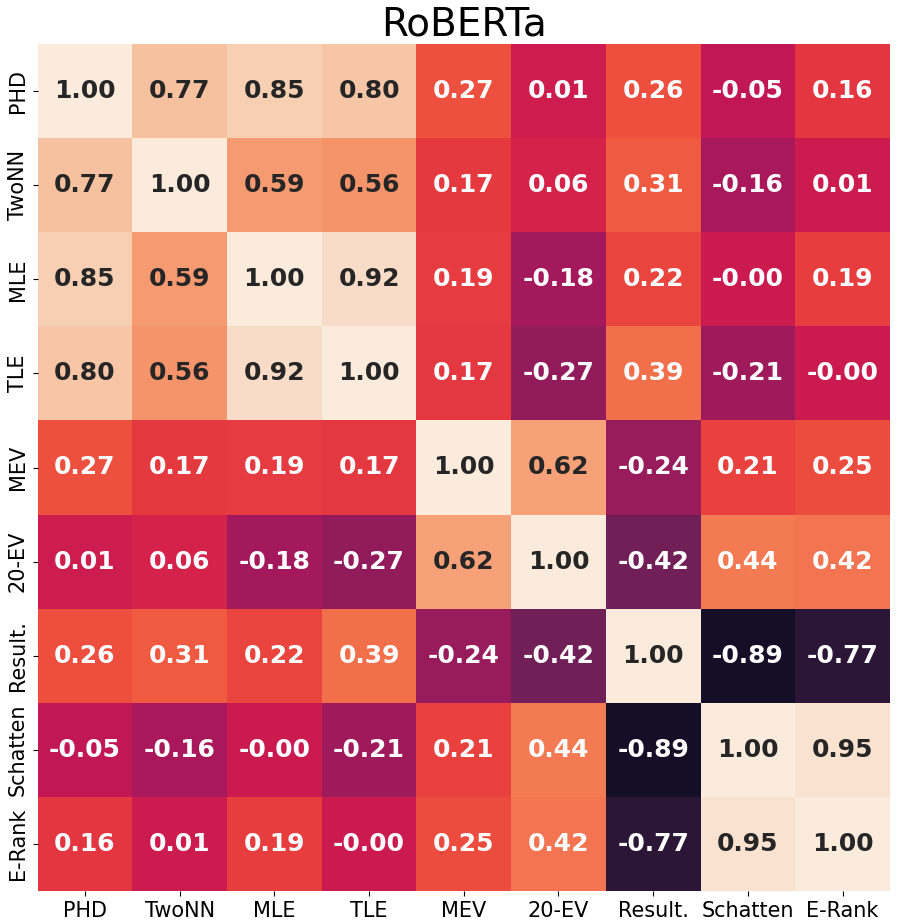} & 
    \includegraphics[width=0.48\linewidth]{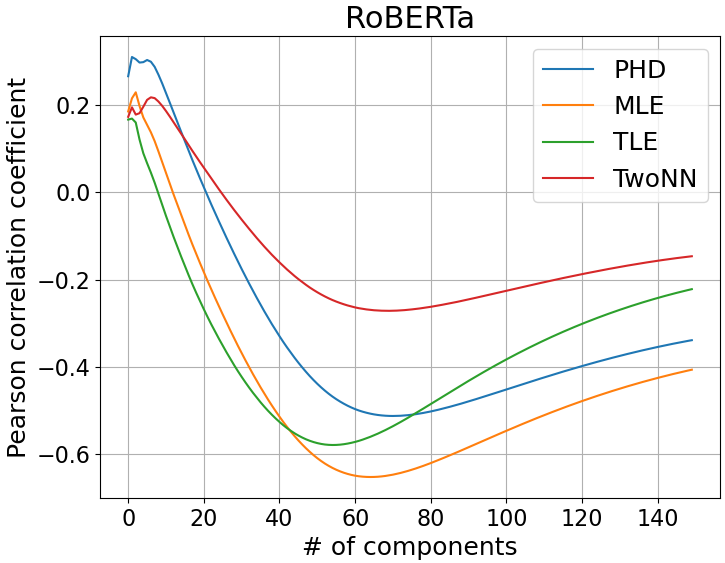}
    \end{tabular}
    
    \caption{Correlation between Intrinsic Dimensionality (ID) and distribution metrics for human text embeddings (left), correlation between ID and cumulative explained variance (right).}
    \label{fig:anisotropy_corr}
\end{figure*}

We calculate the Pearson correlation coefficient between these metrics and 4 intrinsic dimensionality estimators on human-written text data from COLING dataset. Figure~\ref{fig:anisotropy_corr} presents the results for embeddings obtained from RoBERTa-base, Gemma2-2B and Qwen2.5-1.5B models. We can see that different estimators of intrinsic dimensionality (PHD, TwoNN, MLE, and TLE) are highly correlated with each other; for Gemma2 and Qwen2.5 models, there is also strong negative correlation between them and maximal explained variance. The overall picture for RoBERTa is slightly different from other two models, likely because it is an encoder model while Gemma and Qwen are decoders. 

From plots on the right half of the Figure~\ref{fig:anisotropy_corr} we can see that the negative correlation between intrinsic dimensionalities and cumulative explained variance grows with the number of components up to a certain point and then start to decrease. In all cases, the weakest (i.e., closest to zero) correlation is for the TwoNN estimator, which is to be expected because it relies almost entirely on the local structure of the data while principal components are computed for the whole dataset.

\section{Linguistic, syntax and other text properties}

\subsection{PHD dependence on text length (tokens)}\label{app:length}
\begin{figure}[H]
        \centering
        \includegraphics[width=\linewidth]{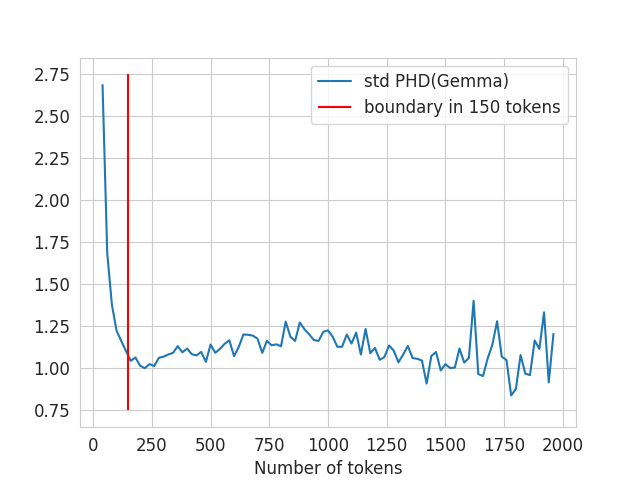}
        \caption{Standard deviation of PHD (Gemma) in 20-token bins.}
        \label{fig:std_phd_gemma}
\end{figure}

\begin{figure}[H]
        \centering
        \includegraphics[width=\linewidth]{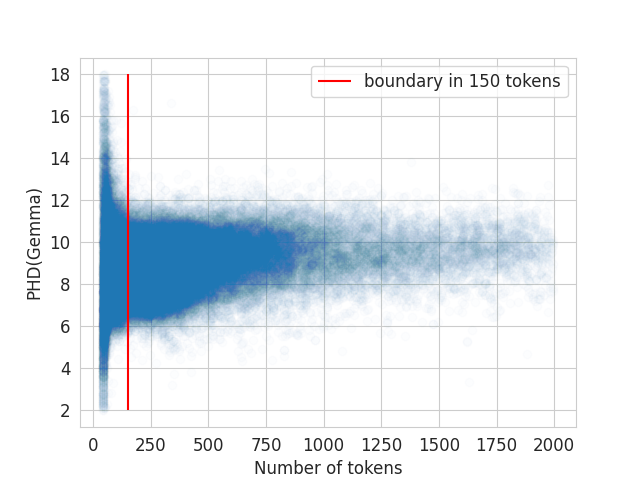}
        \caption{Scatter plot of Text length in tokens and PHD(Gemma)}
        \label{fig:scatter_plot}
\end{figure}

Fig.~\ref{fig:std_phd_gemma} shows the standard deviation of PHD as a function of human text length. The variance is high for short texts and stabilizes beyond about 150 tokens.
In addition, Fig. ~\ref{fig:scatter_plot} shows a large spread of PHD(Gemma) values for short texts.

\subsection{Syntatic diversity}\label{app:syntax_div}
\begin{figure}[H]
        \centering
        \includegraphics[width=\linewidth]{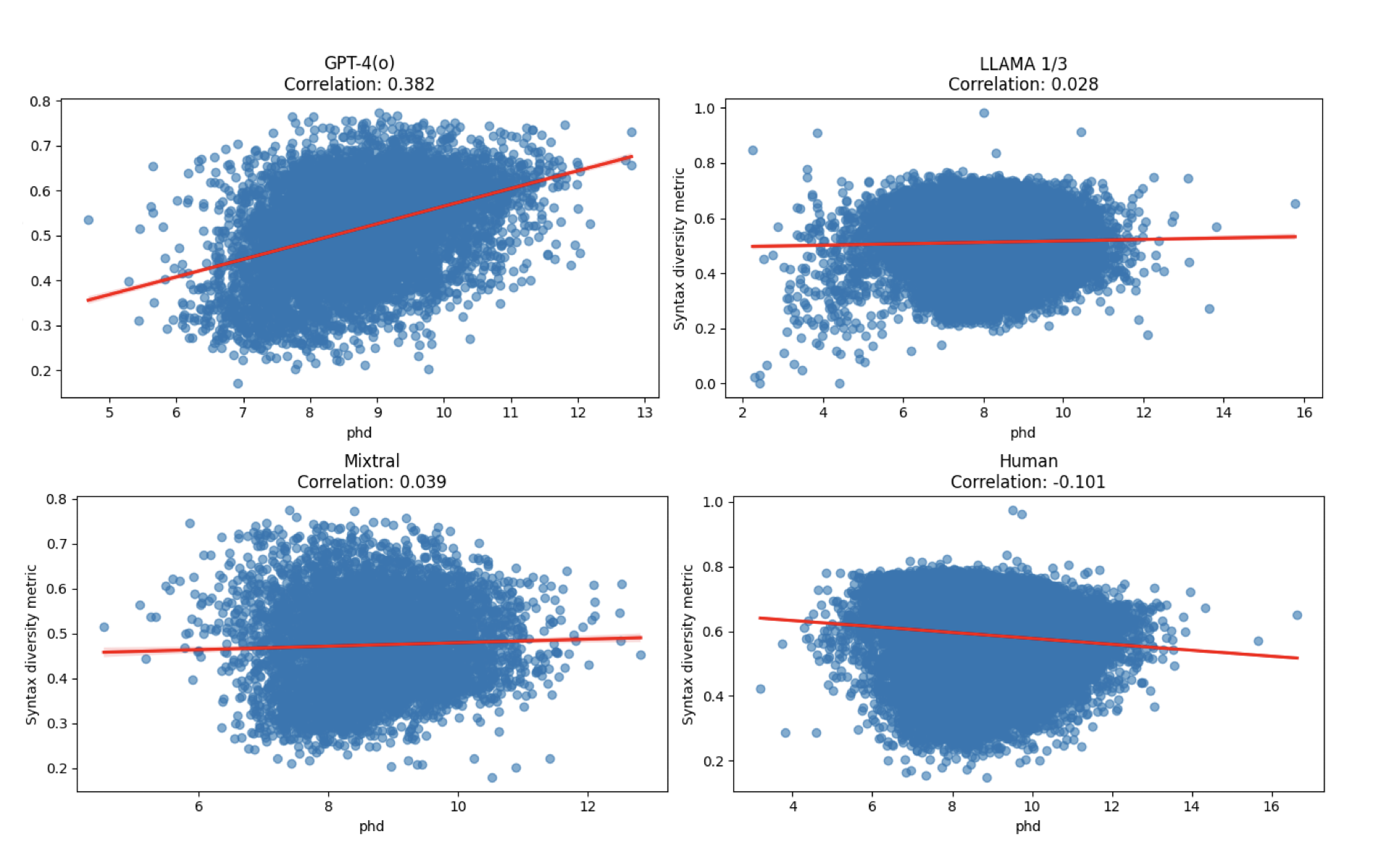}
        \caption{Correlation between syntatic diversity and PHD metric}
        \label{fig:corr_models_phd_syntax}
\end{figure}

We observe that the PHD metric captures little of the syntactic diversity present in texts as in  Fig.~\ref{fig:corr_models_phd_syntax}. Correlation between PHD and syntax diversity and correlation between PHD and POS compression ratio is small and differs among generation LLM models as depicted by Table~\ref{tab:corr_table_models}.

In addition we see that correlation for different sub sources of text: such as arxiv, imbd is diffenrent in sign, but small as it is below forty percent as shown by Table~\ref{tab:corr_table_domains}.
\begin{table}[h]
\centering
\begin{tabular}{|l|r|r|}
\hline
\textbf{Model} & \textbf{CR\_pos} & \textbf{Syntax Diversity} \\
\hline
\rowcolor{blue!30} peerread & 0.235 & \cellcolor{green!50}0.247 \\
\rowcolor{blue!25} cmv & 0.196 & \cellcolor{green!40}0.185 \\
\rowcolor{red!15} outfox & -0.055 & \cellcolor{red!10}-0.016 \\
\rowcolor{blue!25} eli5 & 0.187 & \cellcolor{green!35}0.146 \\
\rowcolor{red!20} wikipedia & -0.086 & \cellcolor{green!25}0.122 \\
\rowcolor{blue!25} wp & 0.167 & \cellcolor{green!50}0.240 \\
\rowcolor{blue!15} reddit & 0.063 & \cellcolor{green!25}0.129 \\
\rowcolor{blue!40} hswag & 0.317 & \cellcolor{green!55}0.267 \\
\rowcolor{blue!15} reddit\_eli5 & 0.044 & \cellcolor{green!35}0.147 \\
\rowcolor{blue!20} finance & 0.155 & \cellcolor{green!45}0.197 \\
\rowcolor{blue!15} xsum & 0.050 & \cellcolor{green!10}0.011 \\
\rowcolor{blue!5} sci\_gen & 0.002 & \cellcolor{red!15}-0.054 \\
\rowcolor{red!15} wikihow & -0.060 & \cellcolor{green!25}0.124 \\
\rowcolor{blue!10} arxiv & 0.036 & \cellcolor{green!20}0.089 \\
\rowcolor{red!10} medicine & -0.037 & \cellcolor{red!10}-0.034 \\
\rowcolor{blue!20} tldr & 0.099 & \cellcolor{green!15}0.043 \\
\rowcolor{red!5} yelp & -0.004 & \cellcolor{red!20}-0.074 \\
\rowcolor{blue!30} roct & 0.223 & \cellcolor{green!15}0.038 \\
\rowcolor{blue!20} squad & 0.139 & \cellcolor{green!20}0.091 \\
\rowcolor{blue!30} cnn & 0.239 & \cellcolor{green!70}0.355 \\
\rowcolor{blue!25} open\_qa & 0.162 & \cellcolor{green!10}0.025 \\
\rowcolor{red!45} imdb & -0.200 & \cellcolor{red!40}-0.164 \\
\rowcolor{blue!5} wiki\_csai & 0.004 & \cellcolor{red!5}-0.012 \\
\rowcolor{red!45} dialogsum & -0.223 & \cellcolor{green!60}0.291 \\
\rowcolor{red!10} pubmed & -0.016 & \cellcolor{red!15}-0.055 \\
\hline
\end{tabular}
\caption{Pearson correlation of CR\_pos and Syntax Diversity with phd dimension by source of text}
\label{tab:corr_table_domains}
\end{table}

\begin{table}[htbp]
\centering
\caption{Model Group Performance Metrics}
\label{tab:model_metrics}
\begin{tabular}{lcc}
\toprule
\textbf{Model Group} & \textbf{CR\_pos} & \textbf{Syntax\_Diversity} \\
\midrule
GPT-4(o)      & \cellcolor{redcell} -0.206 & \cellcolor{greencell} 0.382 \\
LLAMA 1/3     & \cellcolor{yellowcell} 0.201 & \cellcolor{yellowcell} 0.028 \\
Cohere        & \cellcolor{yellowcell} -0.042 & \cellcolor{greencell} 0.190 \\
GPT-3/3.5     & \cellcolor{yellowcell} -0.044 & \cellcolor{yellowcell} 0.185 \\
Mixtral       & \cellcolor{yellowcell} -0.015 & \cellcolor{yellowcell} 0.039 \\
Human         & \cellcolor{yellowcell} 0.030 & \cellcolor{orangecell} -0.101 \\
Bloom/BloomZ  & \cellcolor{yellowcell} 0.074 & \cellcolor{yellowcell} 0.061 \\
Gemma         & \cellcolor{yellowcell} 0.016 & \cellcolor{yellowcell} 0.124 \\
OPT           & \cellcolor{redcell} -0.135 & \cellcolor{yellowcell} 0.170 \\
GPT-J/NeoX    & \cellcolor{redcell} -0.320 & \cellcolor{yellowcell} 0.062 \\
Dolly         & \cellcolor{yellowcell} 0.195 & \cellcolor{orangecell} -0.030 \\
GLM           & \cellcolor{greencell} 0.603 & \cellcolor{yellowcell} 0.414 \\
Flan-T5/T0    & \cellcolor{greencell} 0.744 & \cellcolor{greencell} 0.297 \\
\bottomrule
\end{tabular}
\caption{Pearson correlation of OS compression ratio (CR\_pos) and Syntax Diversity with phd dimension}
\label{tab:corr_table_models}
\end{table}

This is because PHD is largely insensitive to syntax in high-quality texts. Any apparent correlation with syntactic features is likely driven by noisy or low-quality  outputs rather than genuine structural variation. Moreover, domain differences do not significantly affect the relationship between PHD and syntactic diversity or POS compression ratio for good texts. The primary drivers of PHD are, in fact, lexical and thematic diversity, not syntax.

\begin{figure}[H]
        \centering
        \includegraphics[width=\linewidth]{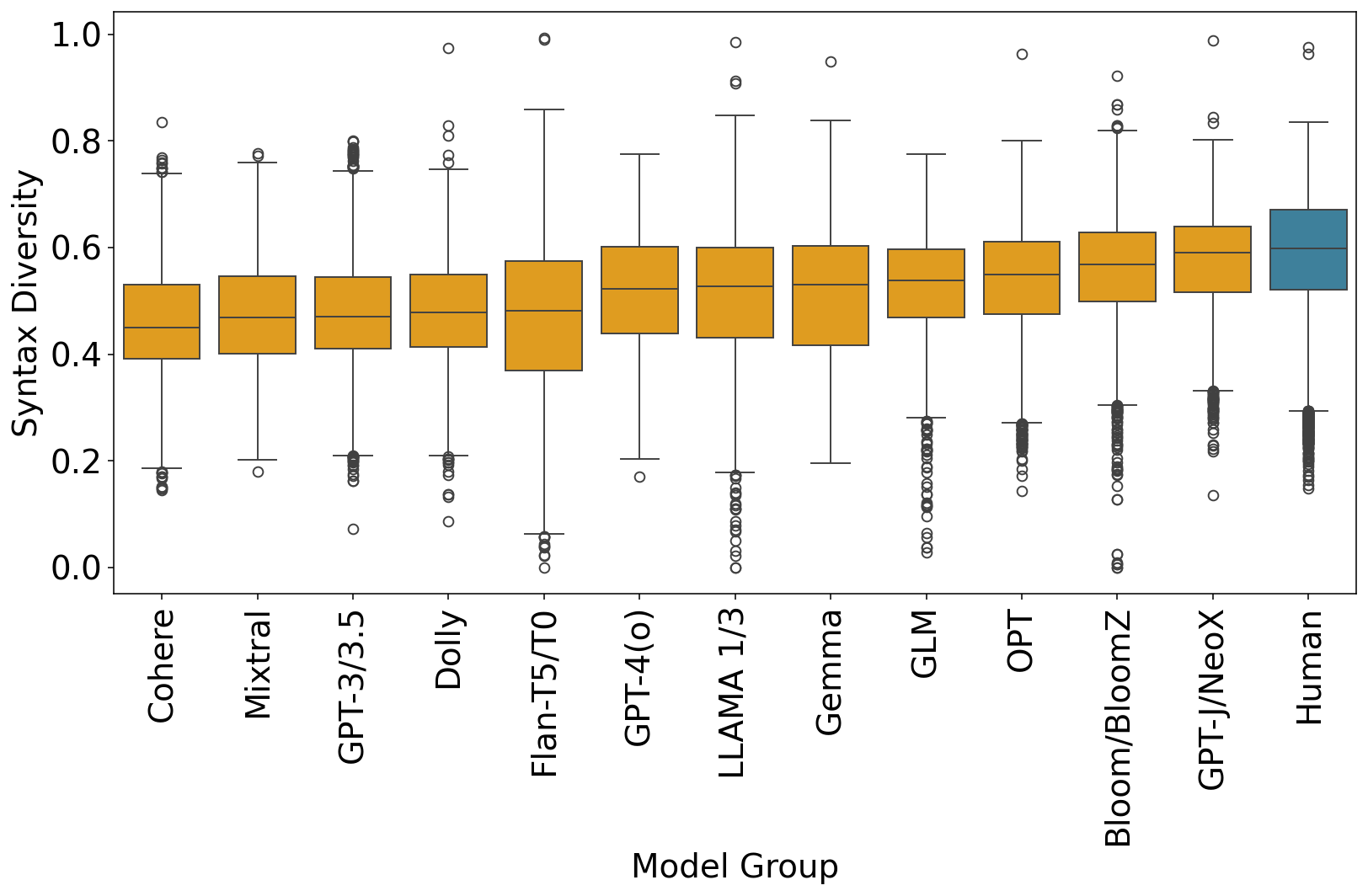}
        \caption{Syntax diversity across text models (human-written and generated by different models)}
        \label{fig:syntax_box_mode}
\end{figure}

In terms of syntactic diversity, human texts are, on average, the most diverse among current families of LLM-generated texts (Fig. ~\ref{fig:syntax_box_mode}), making syntax another factor that should be considered in detecting AI-generated content.

\begin{figure}[H]
        \centering
        \includegraphics[width=\linewidth]{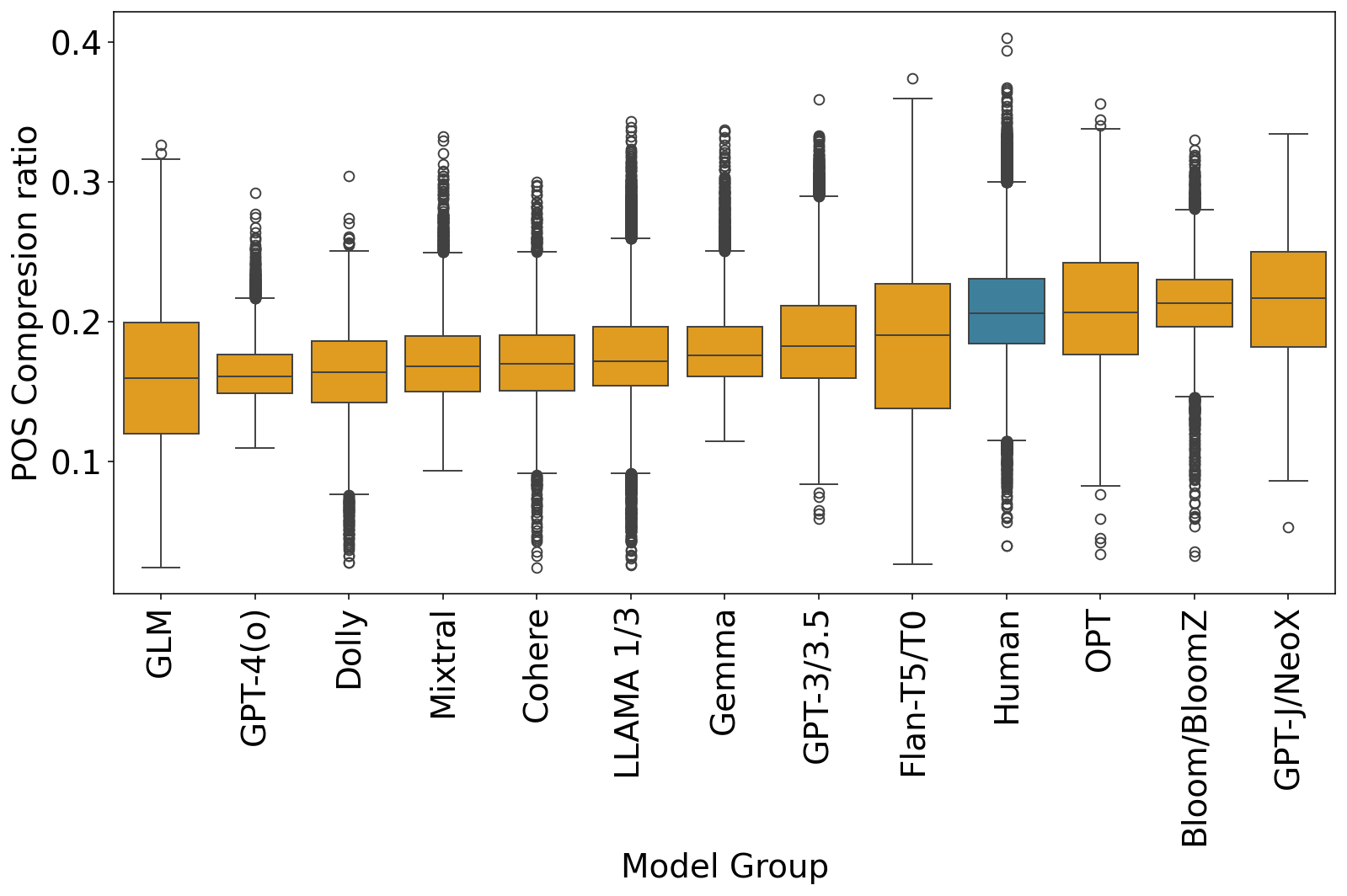}
        \caption{POS compression ratio across text models (human-written and generated by different models)}
        \label{fig:cr_pos_box_model}
\end{figure}

Part-of-speech (POS) tagging compression ratios are relatively high for human-written texts (Fig.~\ref{fig:cr_pos_box_model}), though not the highest among all model groups. However, human texts exhibit notably low variance in this metric, indicating consistent syntactic structure across samples.

Moreover,we categorized the texts by source into three distinct groups based on their PHD values: Low-dimensional, Medium-dimensional, and High-dimensional groups.

The Low-dimensional group, which we designate as "Science \&  Tech", comprises sources such as PubMed, arXiv, Medicine, SciGen, and Wiki -CSAI. This category represents technical and scientific literature from various specialized fields, including medical research and scientific publications.

The Medium-dimensional group, labeled "News \& Info", encompasses news and informational resources including SQuAD, DialogSum, Wikipedia, TLDR, CNN, PeerRead, and XSum. These sources consist primarily of factual reporting, encyclopedia entries, and summarized information.

The High-dimensional group, categorized as "Opinion \& Forum", contains sources such as Outfox, Reddit-ELI5, Reddit, Finance, WikiHow, Yelp, ELI5, IMDb, CMV, and WP. This group represents opinion-based content, including forum discussions, user reviews, crowd-sourced advice, and personal perspectives across various platforms.

\begin{figure}[H]
        \centering
        \includegraphics[width=\linewidth]{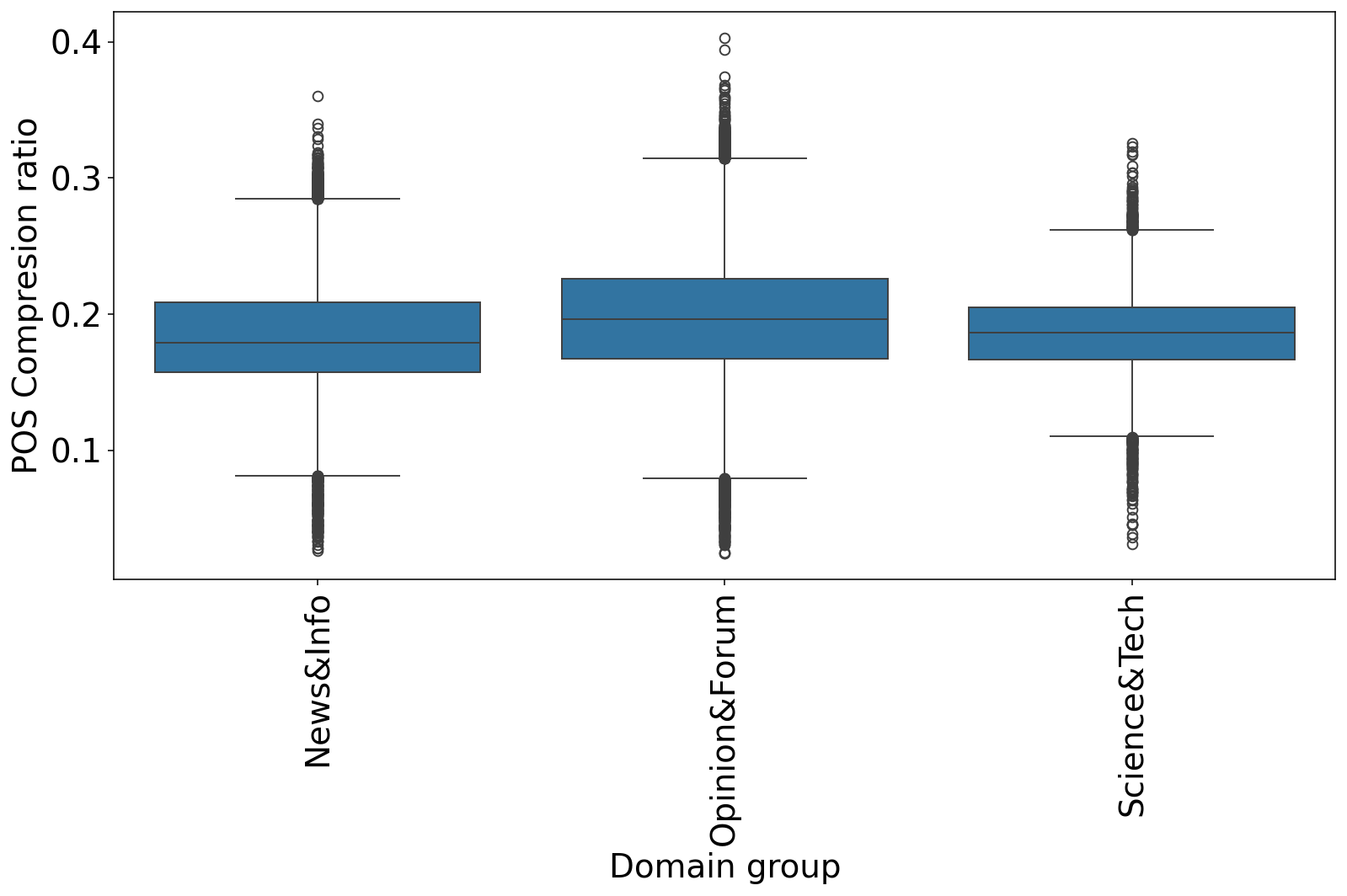}
        \caption{POS Compression ration across text domains}
        \label{fig:cr_box_domain}
\end{figure}

Across these three groups the syntatic diversity metric do not differ a lot, as well as POS Compression ratio Fig.~\ref{fig:cr_box_domain}.

\begin{figure}[H]
        \centering
        \includegraphics[width=\linewidth]{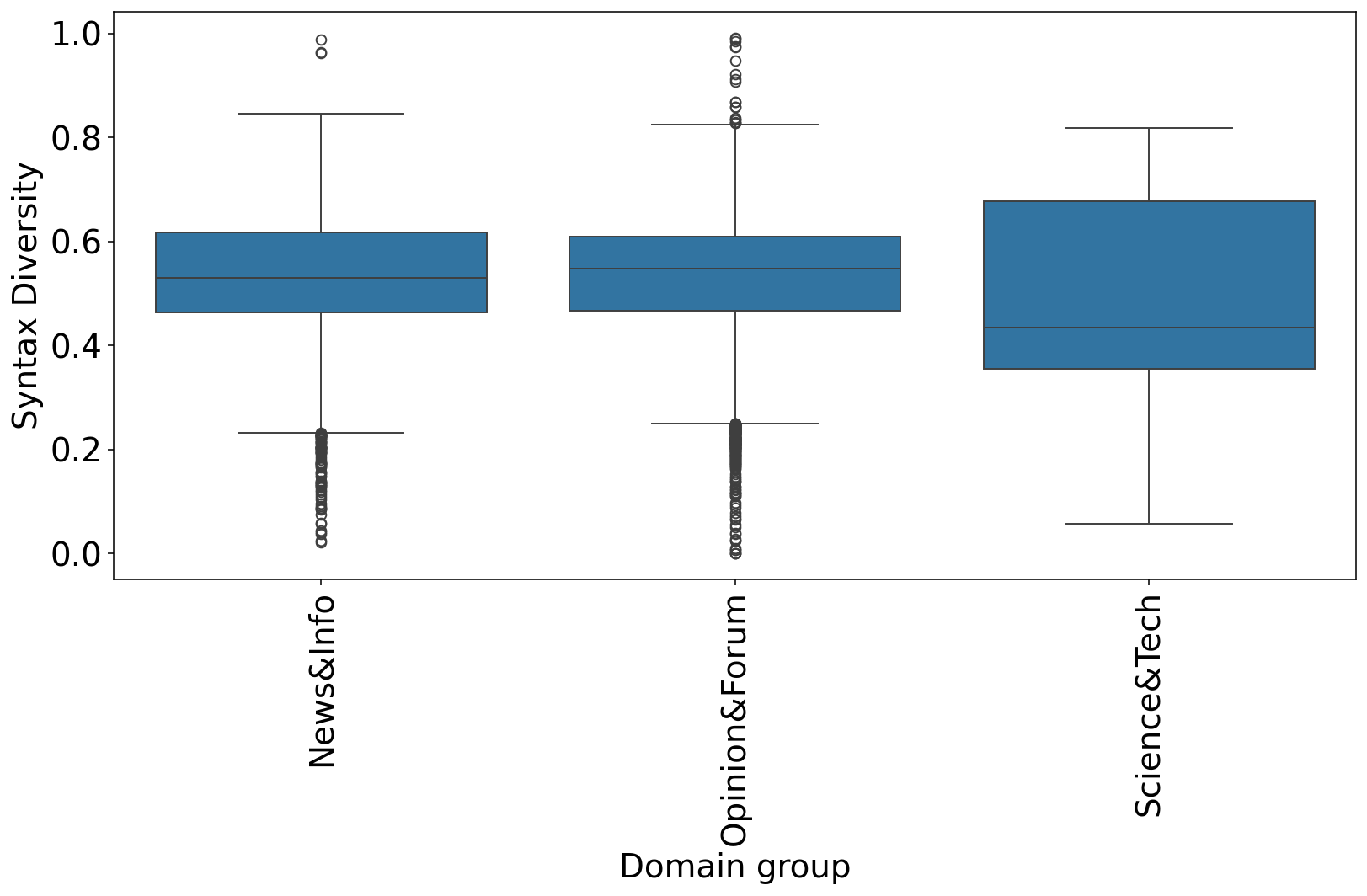}
        \caption{Syntax diversity across text domains}
        \label{fig:syntax_box_domain}
\end{figure}

This can be certainly attributed to generation model itself as distribution of syntax diversity across different domains are similar with insignificantly smaller values for "Science\&Tech" domain as shown by Fig.~\ref{fig:syntax_box_domain}.

\subsection{More details on TAACO features and their relations with ID}
\label{sec:id_taaco}

\begin{figure*}[th!]
        \centering
        \includegraphics[width=0.32\linewidth]{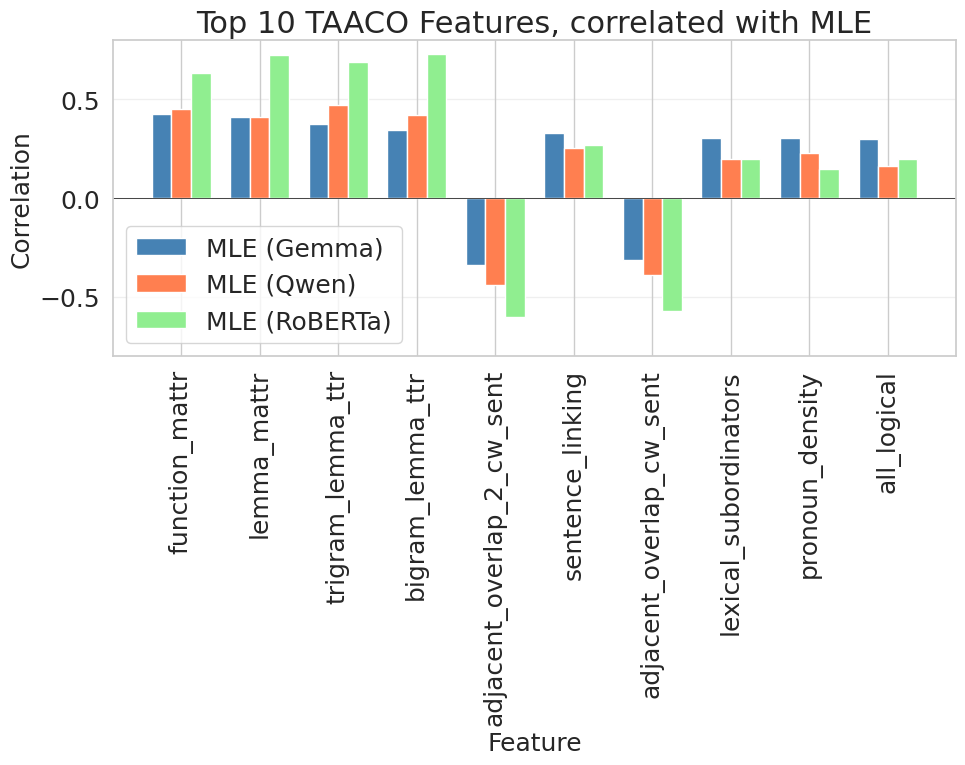}
        \includegraphics[width=0.32\linewidth]{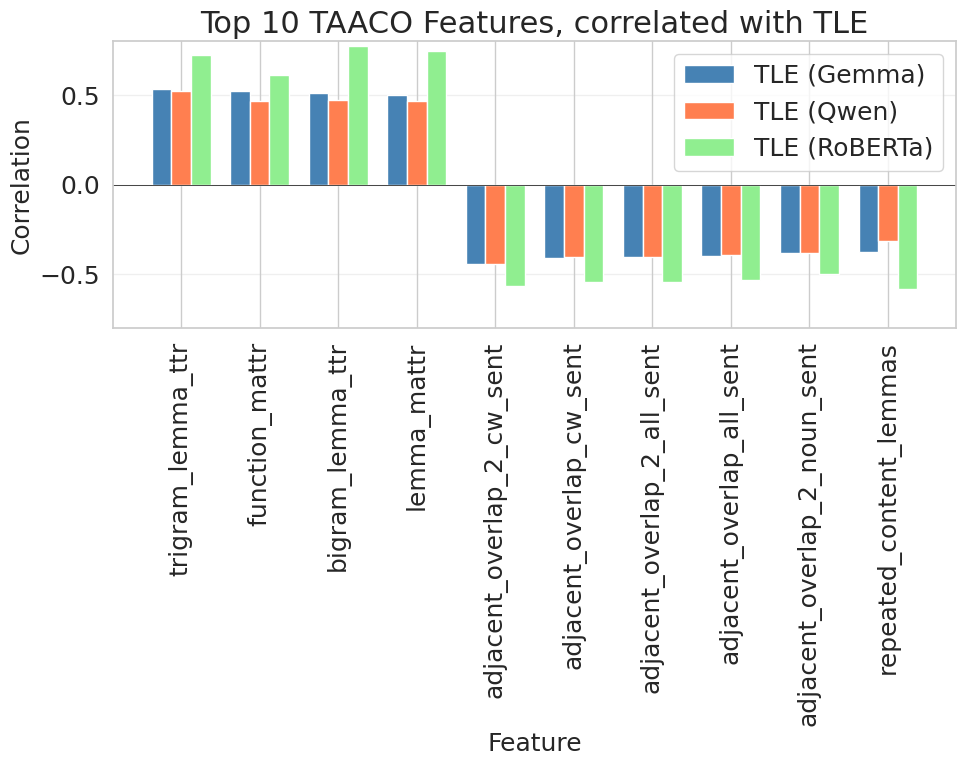}
        \includegraphics[width=0.32\linewidth]{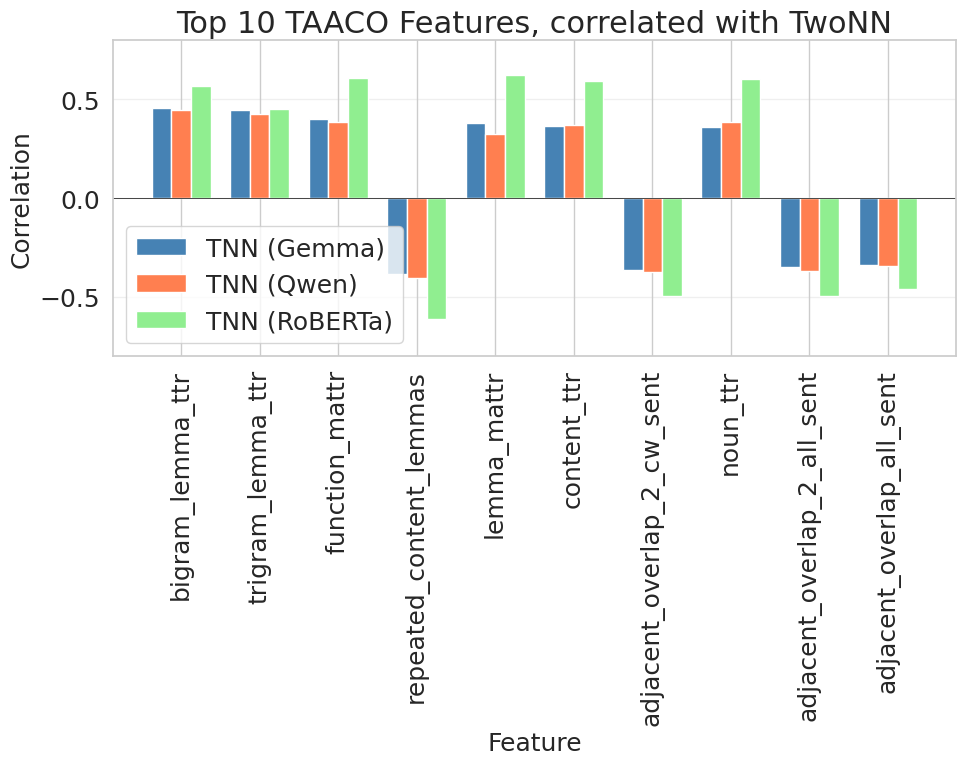}
        \caption{Top-10 features from TAACO with the strongest correlation with MLE, TLE, TwoNN, calculated upon Gemma embeddings, correspondingly. Correlations with ID, calculated upon other embeddings are given as well}
        \label{fig:id_taaco}
\end{figure*}

These are descriptions of the text metrics from TAACO, used on Figures~\ref{fig:id_taaco_corr} and \ref{fig:id_taaco}:

\begin{itemize}
    \item \texttt{function\_mattr} (Moving Average Type-Token Ratio): metric calculates the lexical diversity of function words (articles, prepositions, conjunctions, auxiliary verbs, pronouns) using a sliding window approach to reduce text-length sensitivity. %
    Higher values may indicate varied syntactic structures or register shifts, while lower values suggest more formulaic or consistent grammatical patterns. %
    \item \texttt{lemma\_ttr}, \texttt{bigram\_lemma\_ttr} and \texttt{trigram\_lemma\_ttr}: calculate the type-token ratio of word unigrams, bigrams or trigrams, using their lemmatized forms. %
    Lower values may indicate repetitive words or phrase patterns while higher values suggest more varied lexical combinations.%
    \item \texttt{adjacent\_overlap\_2\_cw\_sent}: measures the overlap of content words (nouns, verbs, adjectives, adverbs) between adjacent sentences, counting multiple occurrences of the same word. %
    Higher values indicate stronger thematic continuity through repeated emphasis on the same content words, %
    while lower values suggest more varied vocabulary between sentences, potentially indicating topic shifts, diverse expression, or weaker local cohesion. %
    \texttt{adjacent\_overlap\_2\_noun\_sent} is very similar, except that only nouns are counted.
    \item \texttt{adjacent\_overlap\_2\_all\_sent} and \texttt{adjacent\_overlap\_all\_sent}: measure overlap of all words between adjacent sentences. The first metric is counting multiple occurrences of repeated words, while the second one is using binary counting (word presence or absence). High values indicate strong repetition and tight cohesion between consecutive sentences, while low values suggest more varied vocabulary and potentially weaker local connections or topic transitions.
    \item \texttt{repeated\_content\_lemmas}: measures the proportion of content word lemmas (nouns, verbs, adjectives, and adverbs in their base forms) that appear more than once throughout the text. Higher values indicate greater lexical cohesion through repetition of meaningful words, while lower values indicate greater lexical diversity but potentially weaker thematic unity or cohesion across the text. %
    \item \texttt{all\_logical}: counts the frequency of logical connectives and discourse markers that express logical relationships in text, including words and phrases like ``therefore'', ``thus'', ``hence'', ``consequently'', ``if-then'', ``because'', and ``since''. Higher values indicate more explicit logical structuring and argumentative discourse, while lower values may characterize more narrative or descriptive texts with implicit rather than explicit logical connections. %
    \item \texttt{sentence\_linking}: counts explicit sentence-linking connectives (e.g., ``moreover'', ``furthermore'', ``however'', ``in addition''). High values indicate explicit discourse structuring with clear transitions between ideas, typical of formal or academic writing, while low values suggest more implicit connections or informal, narrative styles.
    \item \texttt{lexical\_subordinators}: counts subordinating conjunctions that introduce dependent clauses (e.g., ``although'', ``because'', ``while'', ``unless''). High values indicate complex syntactic structures with embedded clauses and sophisticated reasoning, while low values suggest simpler, more direct sentence constructions with coordinate rather than subordinate relationships.
    \item \texttt{pronoun\_density}: measures the proportion of pronouns relative to total words in the text. High values indicate referential cohesion and informal, interactive discourse or narrative style, while low values suggest more explicit naming, formal register, or introductory text where entities are being established rather than referenced.
\end{itemize}

Figure~\ref{fig:id_taaco} and Table~\ref{tab:phd_vs_taaco} show additional correlations between TAACO features and ID estimators upon various models.

\begin{table*}[thbp]
\centering
\resizebox{!}{0.25\textwidth}{
\begin{tabular}{l|ccc|l|cccc}
\toprule
\textbf{TAACO Feature} & \textbf{PHD} & \textbf{PHD} & \textbf{PHD} & \textbf{TAACO Feature} & \textbf{PHD} & \textbf{PHD} & \textbf{PHD} \\
 & (Gemma) & (Qwen) & (RoBERTa) & & (Gemma) & (Qwen) & (RoBERTa)\\
\midrule
bigram\_lemma\_ttr & 0.446 & 0.530 & 0.684 &  adjacent\_overlap\_argument\_sent & -0.227 & -0.306 & -0.447\\
function\_mattr & 0.493 & 0.503 & 0.659 & adjacent\_overlap\_cw\_sent\_div\_seg & -0.270 & -0.328 & -0.370\\
lemma\_mattr & 0.469 & 0.479 & 0.706 & adjacent\_overlap\_2\_verb\_sent & -0.233 & -0.297 & -0.363\\
trigram\_lemma\_ttr & 0.452 & 0.534 & 0.609 & adjacent\_overlap\_binary\_noun\_sent & -0.241 & -0.306 & -0.337\\
adjacent\_overlap\_2\_cw\_sent & -0.404 & -0.486 & -0.600 & adjacent\_overlap\_binary\_2\_noun\_sent & -0.239 & -0.316 & -0.327\\
adjacent\_overlap\_2\_all\_sent & -0.368 & -0.466 & -0.576 & adjacent\_overlap\_adj\_sent & -0.247 & -0.279 & -0.341\\
noun\_ttr & 0.339 & 0.433 & 0.636 &  sentence\_linking & 0.320 & 0.254 & 0.285 \\
repeated\_content\_lemmas & -0.357 & -0.426 & -0.613 &  adjacent\_overlap\_2\_argument\_sent\_div\_seg & -0.227 & -0.295 & -0.336\\
content\_ttr & 0.330 & 0.416 & 0.629 & adjacent\_overlap\_argument\_sent\_div\_seg & -0.213 & -0.272 & -0.331\\
adjacent\_overlap\_cw\_sent & -0.370 & -0.434 & -0.554 & function\_ttr & 0.153 & 0.217 & 0.446  \\
negative\_logical & 0.279 & 0.235 & 0.295 &  adjacent\_overlap\_verb\_sent & -0.219 & -0.262 & -0.333\\
adjacent\_overlap\_all\_sent & -0.358 & -0.438 & -0.545 & all\_temporal & 0.274 & 0.253 & 0.272\\
adjacent\_overlap\_2\_noun\_sent & -0.344 & -0.436 & -0.552 & opposition & 0.279 & 0.232 & 0.287 \\
repeated\_content\_and\_pronoun\_lemmas & -0.274 & -0.365 & -0.565 & determiners & -0.216 & -0.238 & -0.342 \\
adjacent\_overlap\_noun\_sent & -0.309 & -0.387 & -0.508 &  adjacent\_overlap\_binary\_cw\_sent & -0.199 & -0.277 & -0.284 \\
adjacent\_overlap\_2\_fw\_sent & -0.310 & -0.386 & -0.437 & adjacent\_overlap\_all\_sent\_div\_seg & -0.195 & -0.274 & -0.283\\
adj\_ttr & 0.284 & 0.360 & 0.481 & adjacent\_overlap\_2\_adj\_sent\_div\_seg & -0.223 & -0.263 & -0.264 \\
adjacent\_overlap\_fw\_sent & -0.316 & -0.379 & -0.425 & adjacent\_overlap\_binary\_2\_adj\_sent & -0.214 & -0.263 & -0.258 \\
lemma\_ttr & 0.226 & 0.314 & 0.565 & verb\_ttr & 0.127 & 0.216 & 0.375 \\
adjacent\_overlap\_2\_argument\_sent & -0.257 & -0.354 & -0.493 &  adjacent\_overlap\_2\_all\_sent\_div\_seg & -0.173 & -0.264 & -0.259 \\
adjacent\_overlap\_2\_noun\_sent\_div\_seg & -0.298 & -0.351 & -0.382 &  adjacent\_overlap\_adj\_sent\_div\_seg & -0.208 & -0.238 & -0.249\\
adjacent\_overlap\_2\_cw\_sent\_div\_seg & -0.280 & -0.352 & -0.376 & adjacent\_overlap\_binary\_adj\_sent & -0.202 & -0.239 & -0.247 \\
argument\_ttr & 0.182 & 0.284 & 0.522 & adjacent\_overlap\_binary\_2\_cw\_sent & -0.171 & -0.267 & -0.249\\
adjacent\_overlap\_noun\_sent\_div\_seg & -0.282 & -0.328 & -0.375 & all\_negative & 0.259 & 0.191 & 0.217 \\
adjacent\_overlap\_2\_adj\_sent & -0.276 & -0.324 & -0.382 & pronoun\_density & 0.276 & 0.202 & 0.183 \\
\bottomrule
\end{tabular}
}
\caption{Top-50 TAACO features: correlations with PHD across embedding models}
\label{tab:phd_vs_taaco}
\end{table*}

\subsection{Text transformations and their effect on~PHD}\label{app:attack}
We describe three text transformations and examine their effect on the persistent homology dimension across several embedding models.

\begin{table}[h]
\centering
\begin{tabular}{lrrr}
\toprule
 & Gemma & Qwen & Roberta \\
\midrule
PHD & 8.69 & 8.11 & 9.11 \\
PHD Type 1 & 9.47 & 9.83 & 7.91 \\
PHD Type 2 & 8.90 & 8.87 & 6.54 \\
PHD Type 3 & 11.17 & 10.25 & 9.14 \\
\bottomrule
\end{tabular}
\caption{Impact of text transformations on PHD}
\label{tab:diff_embedders}
\end{table}
We consider the following text transformations:
\begin{itemize}
\item Type 1: each letter is independently replaced by a homoglyph with probability $0.2$;
\item Type 2: each letter is replaced by a homoglyph with probability $0.2$, but replacements are applied per word type: if two words are identical before the transformation, they remain identical after;
\item Type 3: the internal letters of each word are shuffled while keeping the first and last letters fixed; identical words remain identical after the transformation.
\end{itemize}
Results for the PHD across transformation types and embedding models are given in  Tab.~\ref{tab:diff_embedders}.

We observe two notable differences between RoBERTa and Gemma/Qwen. First, applying Transformation~3 has little to no effect on the median PHD for RoBERTa, but increases it for Gemma/Qwen. Second, applying Transformations~1 or~2 decreases PHD for RoBERTa, while for Gemma/Qwen these transformations have no effect or slightly increase it. 

This experiment shows that there are differences between RoBERTa and Gemma/Qwen, which are confirmed by Fig ~\ref{fig:id_heatmaps}.

\begin{figure*}[H]
        \centering
        \includegraphics[width=0.99\linewidth]{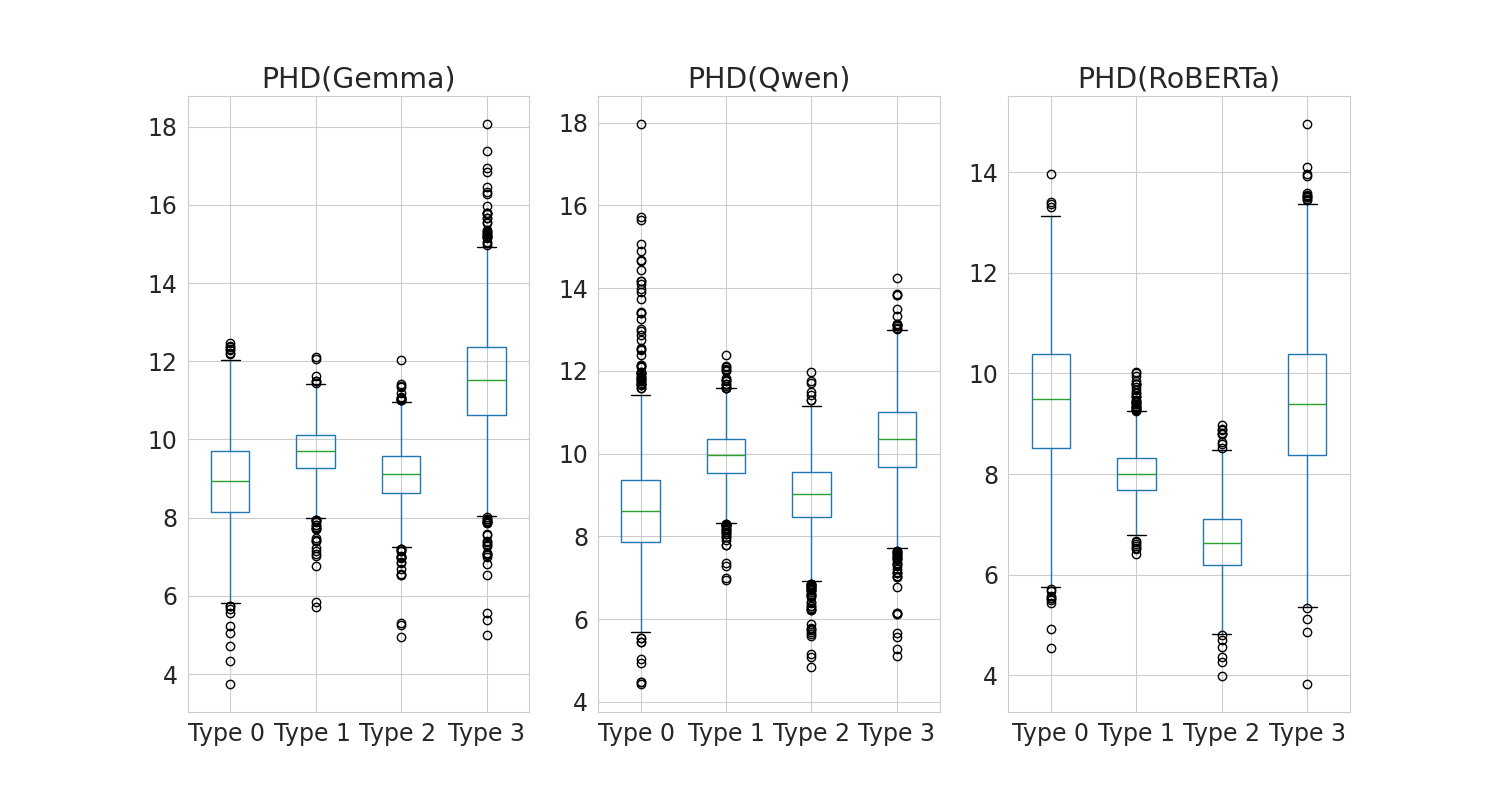}
        \caption{PHD across transformation types for different embeddings}
        \label{fig:diff_embedders}
\end{figure*}

\subsection{Distribution of distances to the nearest neighbor for tokens}
\begin{figure*}[b]
        \centering
        \includegraphics[width=0.8\linewidth]{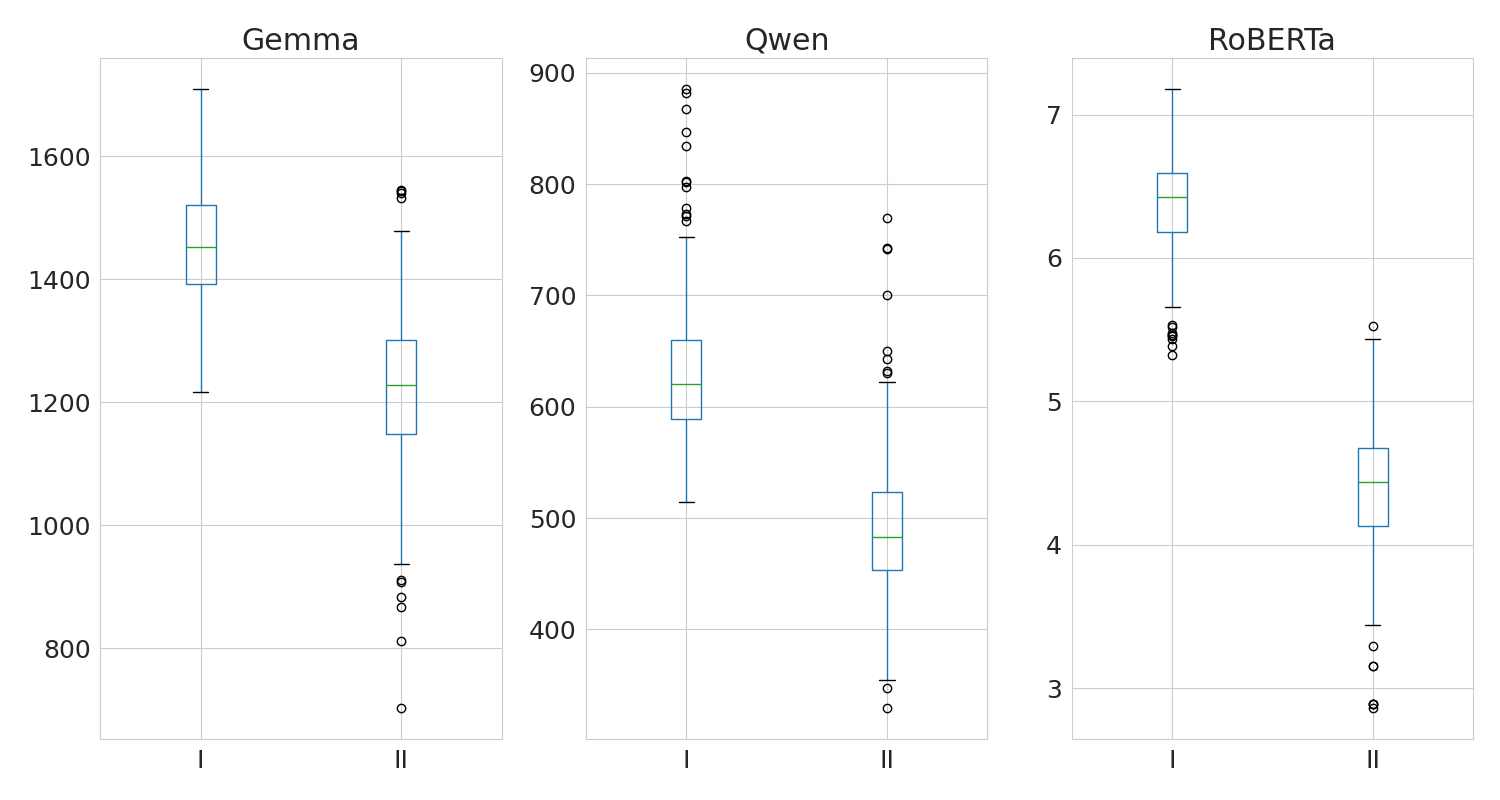}
        \caption{Boxplots of nearest neighbor distances for different types of token}
        \label{fig:phd_diff_sizes2}
\end{figure*} 
To understand why the intrinsic dimensionality of a text positively correlates with its intrinsic dimensionality, we decided to examine the distance between nearest neighbors. To do this, we divided tokens into two types: those with the same nearest neighbor as the token itself (Type I) and those without (Type II). It turned out that the nearest neighbor distances for the Type I tokens are, on average, smaller than those for the Type II, see Fig. ~\ref{fig:phd_diff_sizes2}. This indirectly confirms the positive correlation above. A decrease in the average nearest neighbor distance leads to a decrease in the MST, which in turn may affect its asymptotic behavior.

\subsection{PHD and proficiency level of text}

\begin{figure}[t]
        \centering
        \includegraphics[width=0.99\linewidth]{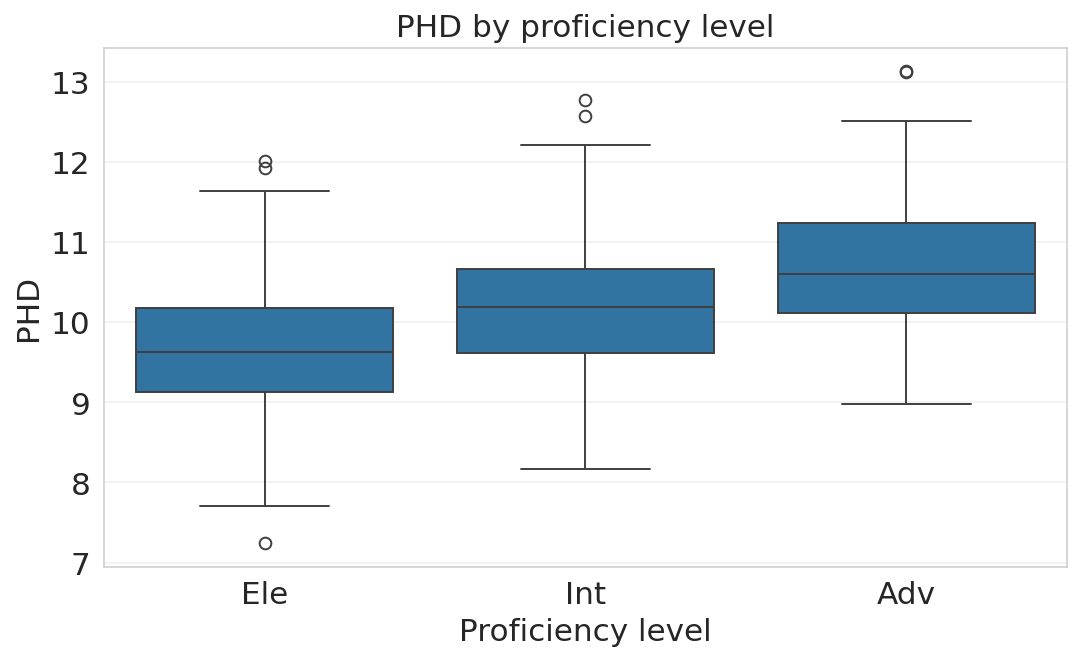}
        \includegraphics[width=0.99\linewidth]{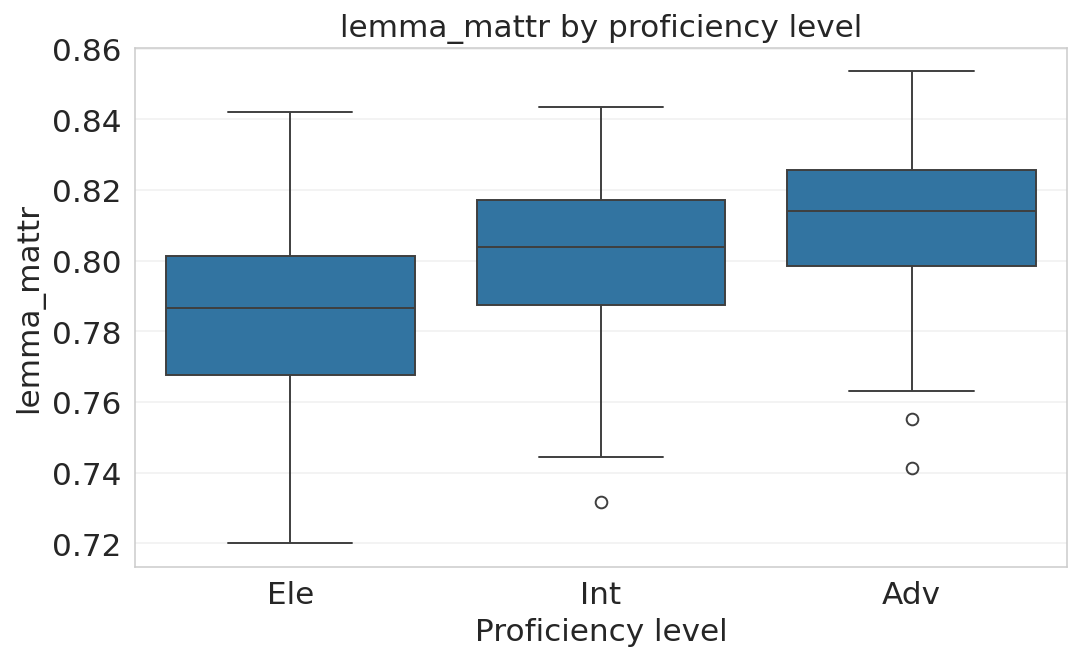}
        \includegraphics[width=0.99\linewidth]{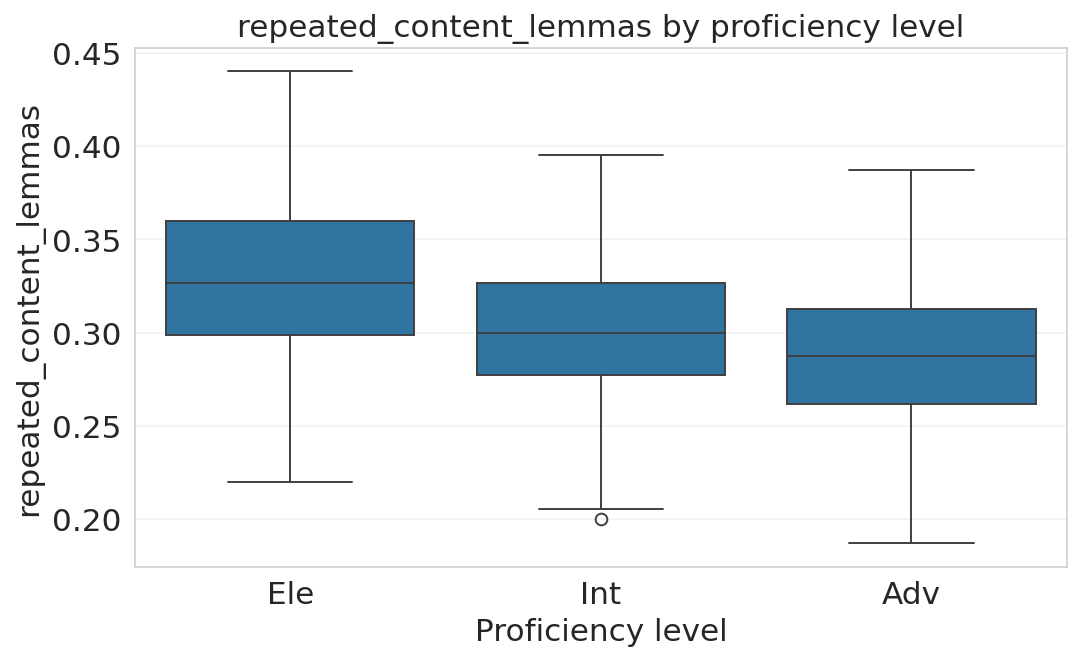}
        \caption{PHD, \texttt{lemma\_mattr} and \texttt{repeated\_content\_lemmas} for texts of various proficiency levels}
        \label{fig:proficiency_phd}
\end{figure}

Figure~\ref{fig:proficiency_phd} shows the distributions of PHD, \texttt{lemma\_mattr}, and \texttt{repeated\_content\_lemmas} for texts written for readers of different proficiency levels. We observe that the PHD of texts written for beginners is lower than that of texts written for intermediate and advanced readers. This can be explained by the lower type–token ratio and higher lemma repetition across consecutive sentences in beginner-level texts.

\subsection{PHD and generation temperature}

We generated 10,000 texts with Qwen-3-8B-base and another 10,000 with Qwen-3-8B-instruct, using 1,000 prompts randomly sampled from the RAID dataset. For each prompt, we generated texts at 10 different temperatures (ranging from 0.2 to 2) and measured their PHD using embeddings from Gemma. The results are shown in the figure.

We observe that the PHD of texts generated by Qwen-3-8B-base increases rapidly as the temperature rises from 0.2 to 0.8, then gradually levels off. In contrast, for Qwen-3-8B-instruct, PHD grows more slowly and almost linearly with temperature. We hypothesize that this difference arises because Qwen-3-8B-base tends to produce many repetitions at low temperatures, while Qwen-3-8B-instruct is less prone to such artifacts and maintains consistent generation quality across all temperatures. The lower panels of the figure support this: low-temperature generations from the base model exhibit very low lexical diversity (measured via \texttt{lemma\_mattr}) and high repetition rates (measured via \texttt{repetitive\_lemmas}), whereas the instruct model does not show this issue.

Figure~\ref{fig:qwen-texts} shows an example of an anomalous text produced by Qwen-3-8B-base and a well-formed text produced by Qwen-3-8B-instruct using the same temperature and prompt. We observe that the base version of Qwen failed to follow the prompt properly and generated many repetitions.

\begin{figure*}[h]
\centering
\begin{minipage}[t]{\linewidth}
\fbox{
\parbox{\linewidth}{
\small
\textbf{Prompt:} \texttt{Write the body of a plot summary for a novel titled "The Drawing of the Dark". Do not give it a title.}
}}
\medskip
\end{minipage}
\begin{minipage}[t]{0.47\linewidth}
\fbox{
\parbox{\linewidth}{
\small
\textbf{Qwen-3-8B-base}\\[3pt]
\texttt{Do not write in the first person. Do not use contractions. Do not use dialogue. Do not use italics. Do not use bold. Do not use hyperlinks. Do not use lists. Do not use tables. Do not use footnotes. Do not use endnotes. Do not use references. Do not use citations. Do not use bibliographies. Do not use indexes. Do not use glossaries. Do not use appendices. Do not use prefaces. Do not use forewords. Do not use introductions. Do not use conclusions. Do not use summaries. Do not use abstracts. Do not use acknowledgments. Do not use dedications. Do not use epigraphs. Do not use prologues. Do not use epilogues. Do not use interludes. Do not use intermissions. Do not use transitions. Do not use flashbacks. Do not use flashforwards. Do not use dream sequences. Do not use fantasy sequences. Do not use hallucinations. Do not use illusions. Do not use visions. Do not use prophecies. Do not use omens. Do not use portents. Do not use auguries. Do not use auguries. Do not use auguries. Do not use auguries. Do not use auguries. Do not use auguries. Do not use auguries. Do not use auguries. Do not use auguries. Do not use auguries. Do not use auguries. Do not use auguries. Do not use auguries. Do not use auguries. Do not use auguries. Do not use auguries. Do not use auguries. Do not use auguries. Do not use auguries. Do not use auguries. Do not use auguries. Do not use auguries. Do not use auguries. Do not use auguries. Do not use auguries. Do not use auguries.}\\
\\
\texttt{...}
}
}
\end{minipage}\hfill
\begin{minipage}[t]{0.47\linewidth}
\fbox{
\parbox{\linewidth}{
\small
\textbf{Qwen-3-8B-instruct}\\[3pt]
\texttt{In a world where the line between reality and illusion is increasingly blurred, a young art student named Elara discovers a mysterious sketchbook hidden in the attic of her late grandmother’s house. The pages are filled with intricate drawings that seem to shift and change when viewed from different angles, revealing hidden symbols and cryptic messages. As Elara begins to study the drawings, she notices that they begin to manifest in the real world—objects appearing in her room, shadows moving on their own, and whispers echoing through the halls of her home.}\\
\texttt{Drawn deeper into the mystery, Elara learns that the sketchbook is a relic from an ancient order of artists known as the Keepers of the Veil, who once protected the world from a force known as the Dark. This force, a sentient void that feeds on fear and uncertainty, was sealed away centuries ago by the Keepers, who used their art to bind it in a place called the Hollow. But the seal is weakening, and the drawings are not just a record of the past—they are a warning.}\\
\texttt{As Elara delves further into the history of the Keepers, she uncovers a prophecy that speaks of a chosen artist who will either restore the balance or unleash the Dark upon the world. With the help of a reclusive scholar named Kael, who has spent his life studying the Keepers’ lost works, Elara begins to understand the true nature of the drawings and the power they hold. She learns that each drawing is a fragment of a larger, unfinished masterpiece that can either reinforce the seal or break it entirely.}\\
\\
\texttt{...}
}
}
\end{minipage}
\caption{Comparison of texts generated by Qwen-3-8B-base and Qwen-3-8B-instruct with the same prompt and temperature (0.2).}
\label{fig:qwen-texts}
\end{figure*}

These observations show that anomalies in PHD distribution may serve as indicators of low-quality text generation.

\begin{figure*}[t]
        \centering
        \includegraphics[width=0.49\linewidth]{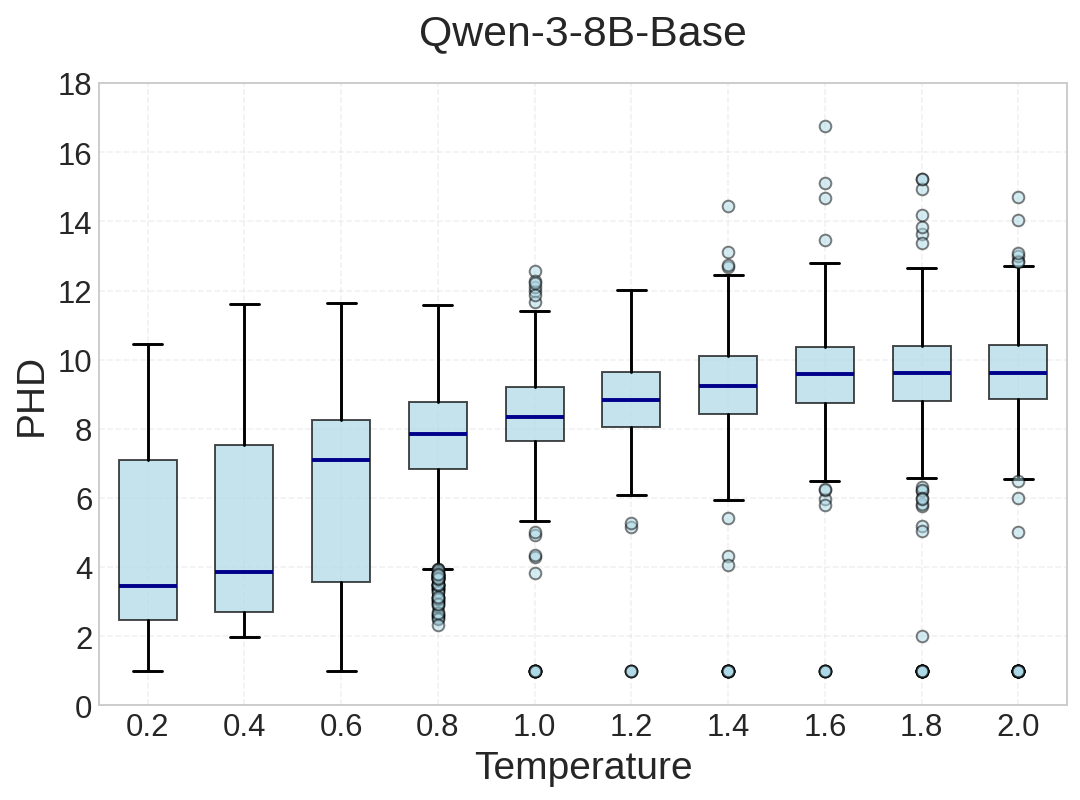}
        \includegraphics[width=0.49\linewidth]{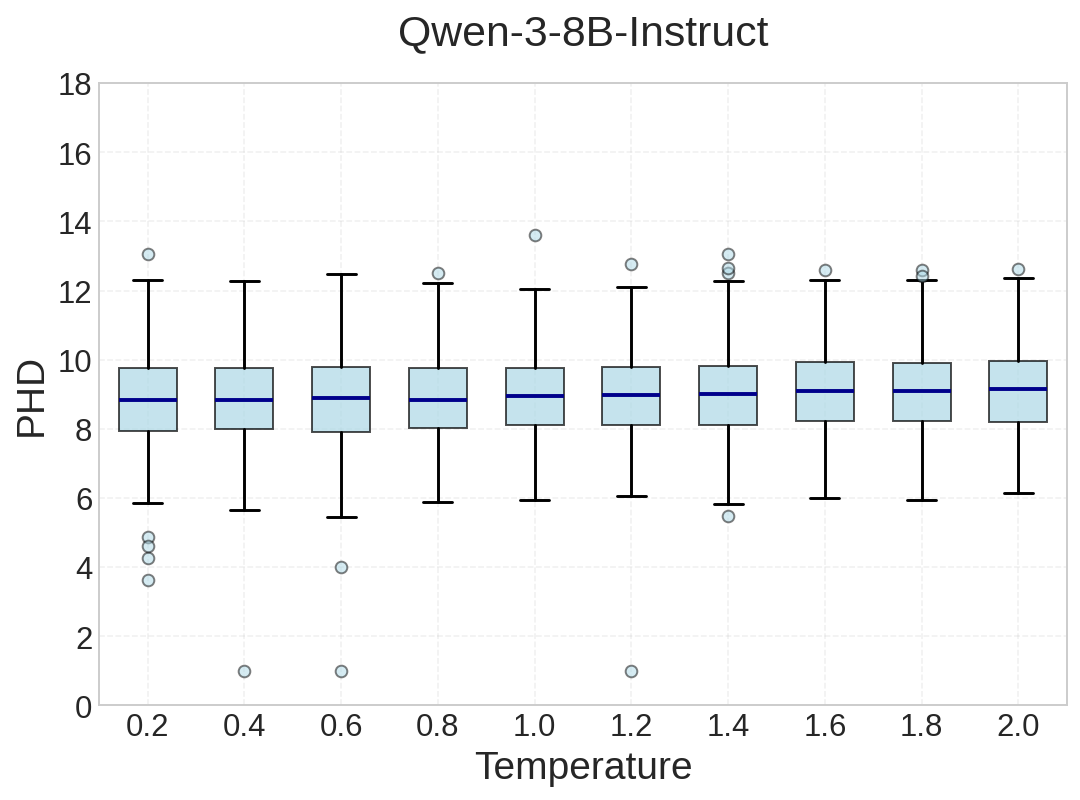}
        \includegraphics[width=0.49\linewidth]{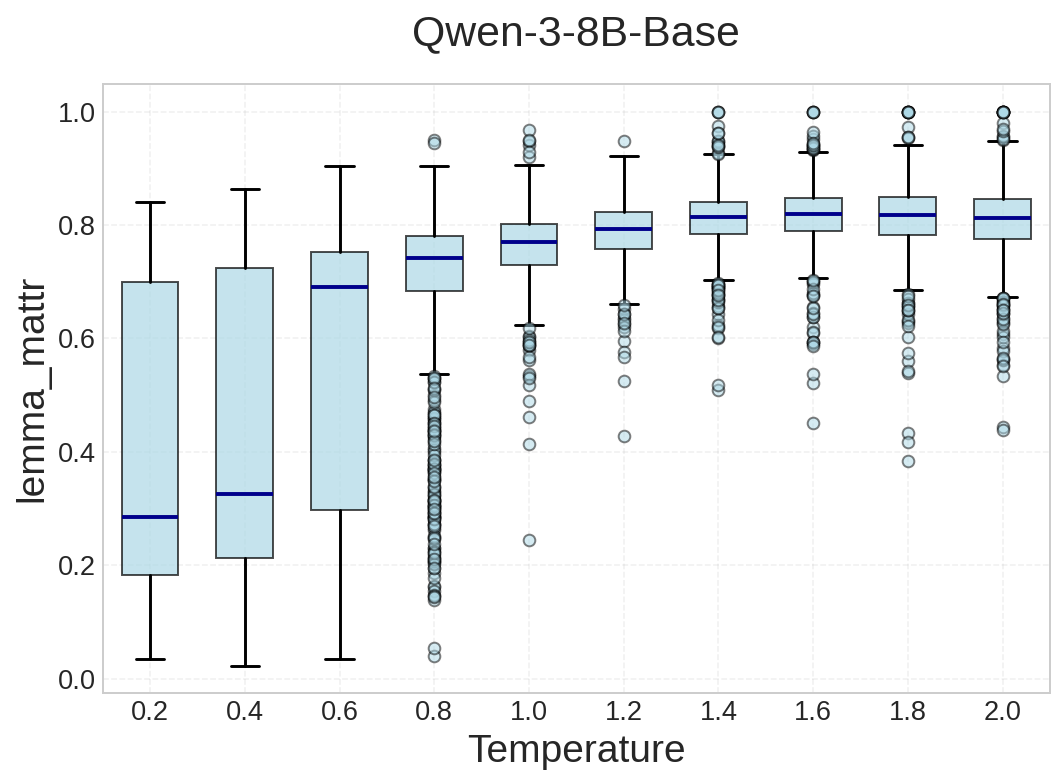}
        \includegraphics[width=0.49\linewidth]{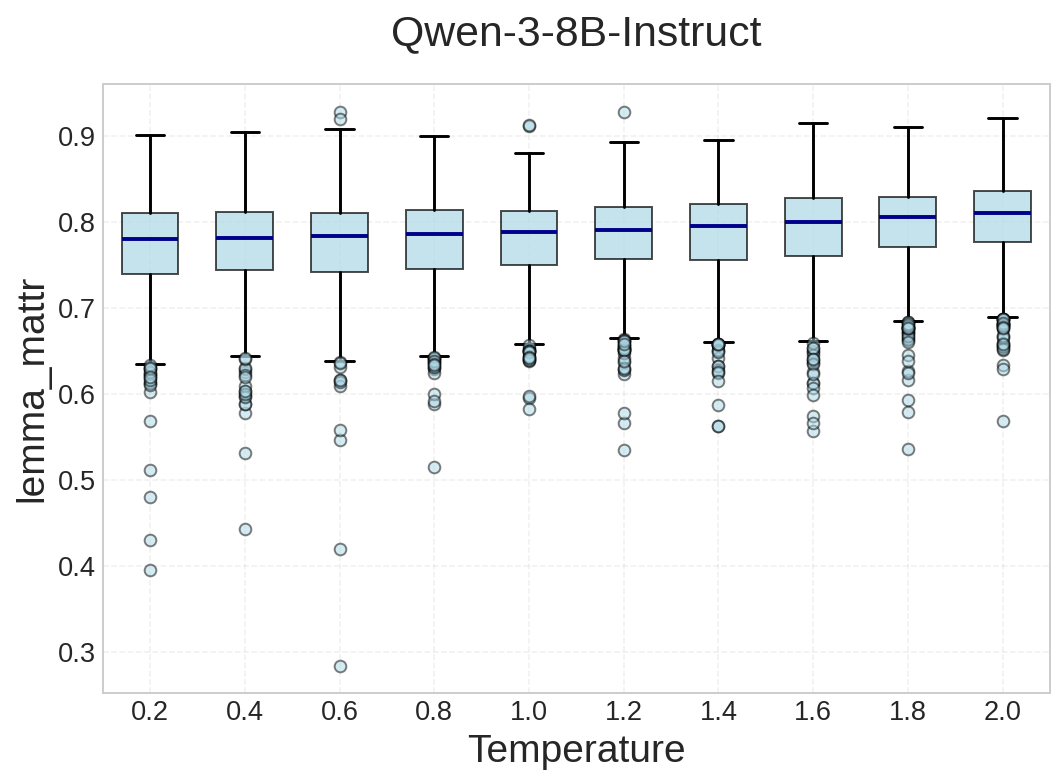}
        \includegraphics[width=0.49\linewidth]{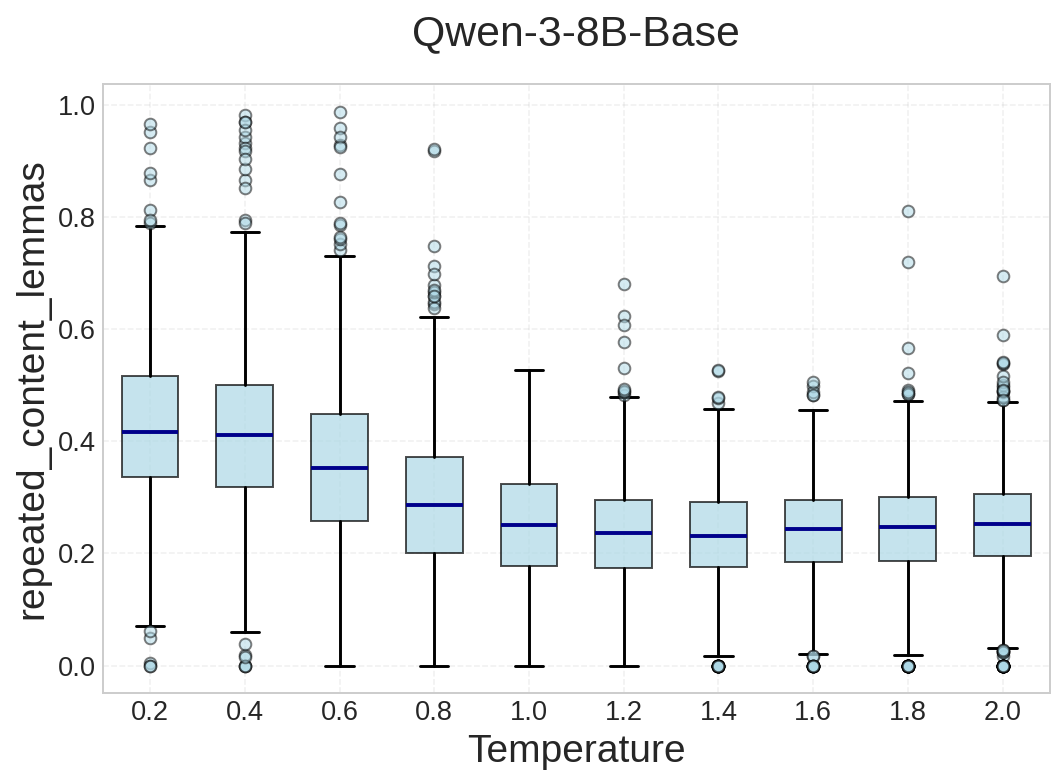}
        \includegraphics[width=0.49\linewidth]{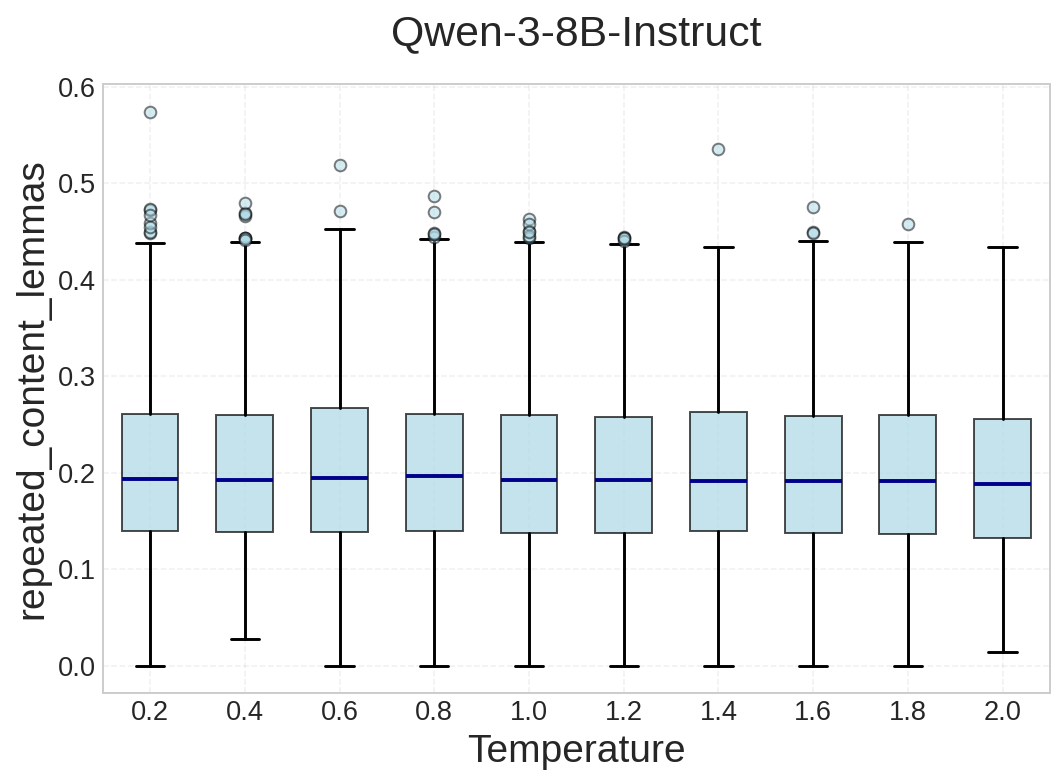}
        \caption{PHD, \texttt{lemma\_mattr} and \texttt{repeated\_content\_lemmas} as a functions of generation temperature}
        \label{fig:qwen_temperature_phd}
\end{figure*} 

\section{SAE-based experiments}

\subsection{Linear approximation for PHD}
We built a simple interpretable model using SAE features, see Tab. ~\ref{tab:feature_interpretations} for details,  and lexical diversity metrics. We trained the model on human texts longer than 150 tokens to predict the PHD(Gemma). We calculated the Pearson and Spearman correlation coefficients for the model's test predictions for various internal dimensionality estimates; see Tab.~\ref{tab:linear_pearson_spearman_coef} for details.
\begin{table}[b!]
\begin{tabular}{lrr}
\toprule
 & pearson & spearman \\
\midrule
PHD(RoBERTa) & 0.597 & 0.643 \\
MLE(RoBERTa) & 0.548 & 0.587 \\
TwoNN(RoBERTa) & 0.512 & 0.556 \\
TLE(RoBERTa) & 0.416 & 0.440 \\
PHD(Gemma) & 0.684 & 0.702 \\
MLE(Gemma) & 0.744 & 0.771 \\
TwoNN(Gemma) & 0.449 & 0.450 \\
TLE(Gemma) & 0.709 & 0.749 \\
PHD(Qwen) & 0.593 & 0.589 \\
MLE(Qwen) & 0.680 & 0.682 \\
TwoNN(Qwen) & 0.430 & 0.402 \\
TLE(Qwen) & 0.706 & 0.710 \\
\bottomrule
\end{tabular}
\caption{Pearson and Spearman correlations for linear model predictions on a test for various estimates of the internal dimension.}
\label{tab:linear_pearson_spearman_coef}
\end{table}

Also, to obtain information about the importance of features, we built linear regressions on the same features, each of which was trained to predict its own estimate of the internal dimension, calculated its importance for each feature, and calculated the rank in such a way that the lower the rank, the more important the feature. Then, for each feature, we took the median of its ranks and got the following Tab. ~\ref{tab:linear_pearson_spearman_coef}. Here MTLD \cite{mtld-paper} and Summer \cite{somers1966statistical} are metrics of lexical diversity which are calculated using the Python library \textit{LexicalRichness} \cite{lexical-richness-python-package}.

\begin{table}[b!]
\begin{tabular}{lll}
\hline
Layer & Feature & \begin{tabular}[c]{@{}l@{}}Median\\    rank\end{tabular} \\ \hline
-     & Summer  & 1                                                        \\
-     & MTLD    & 2                                                        \\
16    & 1693    & 3                                                        \\
16    & 9868    & 3.5                                                      \\
25    & 14085   & 6                                                        \\
16    & 5228    & 7                                                        \\
16    & 14885   & 8                                                        \\
16    & 8104    & 9                                                        \\
16    & 2409    & 9                                                        \\ \hline
\end{tabular}
\caption{Median rank of importance of each feature for the linear approximation model PHD}
\label{tab:feautre_importance}
\end{table}

\subsection{Steering examples}\label{app:steering}
Table~\ref{tab:more_steering_examples} demonstrates the steering effect of the selected features by illustrative examples.

\begin{table*}[ht]
\footnotesize
\begin{tabular}{p{1.3cm}p{0.3cm}p{3.5cm}p{9.2cm}}
\hline
\textbf{Feature} & \textbf{Sign} & \textbf{Shift} & \textbf{Example change (steered)} \\
\hline
\multicolumn{4}{p{15.3cm}}{
\textbf{Default generation (excerpt):} 
\emph{The internet is awash with advice for anyone struggling to return to education after a long break. \dots{}
The first step is to acknowledge the gap in your education and accept that it's okay to feel nervous. Local colleges often offer adult courses to help you catch up.}
}\\[1mm]
24-15879 & -- & BBC-style templated explainer & ``‘I'm not sure what I'm doing with my life,’ she said — the anecdotal lede introducing a structured explainer: paragraph 1 (anecdote), paragraph 2 (context and stats), paragraph 3 (options for learners), paragraph 4 (call to action).'' \\
16-8104 & -- & Social-science report & ``A UK survey found that 12\% of adults had never attended a formal school or college.'' \\
24-4610 & + & Social-media explainer & ``\textbf{Feeling lost?} You're not alone. \textbf{1.~Assess your goals} \textbf{2.~Research options} \textbf{3.~Talk to someone}.'' \\
16-6978 & + & Informal helper tone & ``It’s okay — let’s take it step by step and outline a plan that fits your life.'' \\
\hline
\multicolumn{4}{p{15.3cm}}{
\textbf{Default generation (excerpt):} 
\emph{The awkward silence hung heavy in the air as Sarah and I sat across from each other at the cafe. We were supposed to be discussing her art, but all I could think about was the way her eyes lingered on me.}
}\\[1mm]
16-2433 & -- & Outline / topic-sentence scaffolding & ``The story focuses on: \textbullet~context of the situation \textbullet~comparison with other cases.'' \\
16-5159 & -- & Analytical / detached commentary & ``The conversation is analyzed through a framework of social and cultural factors influencing interaction.'' \\
16-5228 & + & Emotional uncertainty; introspective tone & ``I’m trying to move on, but it’s hard — I still don’t know if she likes me as a friend or more.'' \\
15275 & + & Character-focused vividness; narrative expansion & ``Emily’s fingers trembled as she met Sarah’s gaze, unsure if the warmth in her smile meant love or fear.'' \\
\hline
\multicolumn{4}{p{15.3cm}}{
\textbf{Default generation (excerpt):} 
\emph{Elias Thorne, a brilliant but troubled young programmer, is caught in the middle of this clash of worlds. He's a man of contradictions, haunted by the ghosts of his own making.}
}\\[1mm]
16-5984 & -- & Academic analytical tone & ``The novel's narrative structure is based on competing architectural firms’ design histories within the 1980s urban landscape.'' \\
16-14485 & -- & Expository / research-like narrative & ``The narrative emphasizes the city's socio-economic contrasts and their impact on collective memory.'' \\
15275 & + & Richer plot and personal emotion & ``Our protagonist, Alex, a struggling journalist, fights to reclaim his place in a city that forgot his name.'' \\
16-5228 & + & Psychological depth / uncertainty & ``Anya must race against time, yet her own doubts and past failures threaten to destroy her resolve.'' \\
16-6978 & + & Informal helper tone & ``It’s okay — let’s take it step by step and outline a plan that fits your life.'' \\
\hline
\multicolumn{4}{p{15.3cm}}{
\textbf{Default generation (excerpt):} 
\emph{This paper explores the theoretical framework for unconventional superconductivity in strongly correlated systems, focusing on the interplay between real-space pairing and mean-field theory.}
}\\[1mm]
16-5984 & -- & Analytical / formalized research tone & ``The study reveals the critical role of the fuel supply system, including the pump, lines, and pressure regulator, in performance metrics.'' \\
16-5159 & -- & Repetitive academic phrasing; self-referential & ``This paper examines the causes of this problem and provides a diagnostic framework; this paper is intended to be a guide.'' \\
16-6978 & + & Informal / conversational rewrite & ``Here's the abstract — I’ve kept it short and jargon-free so it’s easier to follow.'' \\
16-9868 & + & Media / popular-science tone & ``The emergence of superconductivity in correlated systems has fascinated scientists for decades — and the story is still unfolding.'' \\
\hline
\end{tabular}
\caption{Steering effect of selected features for four different prompts. ``Sign'' denotes positive or negative correlation of the feature with ID.}\label{tab:more_steering_examples}
\end{table*}

\section{Speeding up the running time of the PHDim calculation.}
The most expensive part of the calculation
PHDim is a calculation of pairwise distances. For long
texts with a length of more than 1024 tokens, this
may take nearly a week on  Intel(R) Core(TM) i7-14700K processor for full COLING-dev dataset. Therefore, we rewrote the code
for calculating the matrix of pairwise distances with
CPU to GPU, which significantly accelerated the work with long texts. For shorter texts, this gives a slight speedup, which is associated with a long transfer of data between the CPU and GPU. The code is located here \footnote[0]{https://anonymous.4open.science/status/TopAnDat-FE46}.f

\section{Computational resources}

For our experiments we used servers with the following computational resources:
\begin{itemize}
    \item 2 x NVIDIA GeForce RTX 4090 GPU + Intel(R) Core(TM) i7-14700K CPU
    \item 4 x NVIDIA A100 80 Gb + Intel(R) Xeon(R) Gold 6348 CPU @ 2.60GHz
\end{itemize}

\section{Artifact licenses}
\begin{itemize}
    \item COLING dataset - The MIT License,  %
    \item RAID dataset - The MIT license, %
    \item TAACO - Creative Commons Attribution-NonCommercial-Sharealike license (4.0),
    \item OneStopEnglish -  CC by-SA 4.0
\end{itemize}

\end{document}

%% file: math_commands.tex
\usepackage{amsmath,amsfonts,bm}

\def\eqref#1{equation~\ref{#1}}

\def\1{\bm{1}}

\DeclareMathAlphabet{\mathsfit}{\encodingdefault}{\sfdefault}{m}{sl}
\SetMathAlphabet{\mathsfit}{bold}{\encodingdefault}{\sfdefault}{bx}{n}